\documentclass[10pt,journal,letterpaper,compsoc]{IEEEtran}
\usepackage{xspace}
\usepackage{times}
\usepackage{epsfig}
\usepackage{graphicx}
\usepackage{amsmath}
\usepackage{amssymb}
\usepackage{dsfont}
\usepackage{multirow}
\usepackage[pagebackref=true,breaklinks=true,letterpaper=true,colorlinks,bookmarks=false]{hyperref}
\usepackage{subfigure}
\usepackage{lscape}
\usepackage{multirow}
\usepackage{algpseudocode}
\usepackage{algorithm}

\def\a{\alpha}

\def\D{\Delta}

\makeatletter
\DeclareRobustCommand\onedot{\futurelet\@let@token\@onedot}
\def\@onedot{\ifx\@let@token.\else.\null\fi\xspace}

\def\eg{\emph{e.g}\onedot} 
\def\ie{\emph{i.e}\onedot} 
 
 \def\vs{\emph{vs}\onedot}
 
\def\etal{\emph{et al}\onedot}
\makeatother

\usepackage{framed}
\usepackage{xcolor}
\definecolor{gainsboro}{RGB}{220,220,220}

\newcommand{\x}{\mathbf{x}}

\newcommand{\X}{\mathcal{X}}
\newcommand{\Y}{\mathcal{Y}}

\def\a{\alpha}
\def\b{\beta}

\def\t{\theta}

\def\p{\varphi}
\def\x2{\chi^2}

\def\D{\Delta}
\def\Re{\mathbb R}

\def\X{{\cal X}}
\def\Y{{\cal Y}}
\def\tX{\tilde{\X}}
\def\tY{\tilde{\Y}}
\def\1{\mathds{1}}

\newcommand{\specialcell}[2][c]{\begin{tabular}[#1]{@{}c@{}}#2\end{tabular}}

\begin{document}

\title{Label-Embedding for Image Classification}

\author{Zeynep~Akata,~\IEEEmembership{Member,~IEEE,}
  Florent~Perronnin,~\IEEEmembership{Member,~IEEE,}
  Zaid~Harchaoui,~\IEEEmembership{Member,~IEEE,}
  and~Cordelia~Schmid,~\IEEEmembership{Fellow,~IEEE}
\IEEEcompsocitemizethanks{\IEEEcompsocthanksitem Z. Akata is currently with the Computer Vision and Multimodal Computing group 
of the Max-Planck Institute for Informatics, Saarbrucken, Germany.
The vast majority of this work was done while Z. Akata was jointly with the Computer Vision group of the Xerox Research Centre Europe and the LEAR group of INRIA Grenoble Grenoble Rh\^one-Alpes.\protect\\
\IEEEcompsocthanksitem F. Perronnin is currently with Facebook AI Research. The vast majority of this work was done while F. Perronnin was with the Computer Vision group of the Xerox Research Centre Europe, Meylan, France.\protect\\
\IEEEcompsocthanksitem Z. Harchaoui and C. Schmid are with the LEAR group of INRIA Grenoble Rh\^one-Alpes, Montbonnot, France.}
}

\IEEEcompsoctitleabstractindextext{%
\begin{abstract}
Attributes act as intermediate representations that enable parameter sharing between classes, a must when training data is scarce. We propose to view attribute-based image classification as a label-embedding problem: each class is embedded in the space of attribute vectors. We introduce a function that measures the compatibility between an image and a label embedding. The parameters of this function are learned on a training set of labeled samples to ensure that, given an image, the correct classes rank higher than the incorrect ones. Results on the Animals With Attributes and Caltech-UCSD-Birds datasets show that the proposed framework outperforms the standard Direct Attribute Prediction baseline in a zero-shot learning scenario. Label embedding enjoys a built-in ability to leverage alternative sources of information instead of or  in addition to attributes, such as \eg class hierarchies or textual descriptions. Moreover, label embedding encompasses the whole range of learning settings from zero-shot learning to regular learning with a large number of labeled examples.
\end{abstract}
}

\maketitle

\section{Introduction}
We consider the image classification problem where the task is to annotate a given image with one (or multiple) class label(s) describing its visual content. Image classification is a prediction task: the goal is to learn from a labeled training set a function 
$f: \X \rightarrow \Y$ which maps an input $x$ in the space of images $\X$ to an output $y$ in the space of class labels $\Y$. In this work, we are especially interested in the case where classes are related (\eg they all correspond to animals), but where we do not have {\em any (positive) labeled sample} for some of the classes. This problem is generally referred to as zero-shot learning \cite{FEHF09,LNH09,LEB08,PPH09}. Given the impossibility to collect labeled training samples in an exhaustive manner for all possible visual concepts, zero-shot learning is a problem of high practical value. 

An elegant solution to zero-shot learning, called attribute-based learning, has recently gained popularity in computer vision. Attribute-based learning consists in introducing an intermediate space ${\cal A}$ referred to as {\em attribute} layer \cite{FEHF09,LNH09}. 
Attributes correspond to high-level properties of the objects which are {\em shared} across multiple classes, which can be detected by machines and which can be understood by humans. Each class can be represented as a vector of class-attribute associations according to the presence or absence of each attribute for that class. Such class-attribute associations are often binary. As an example, if the classes correspond to animals, possible attributes include ``has paws'', ``has stripes'' or ``is black''.
For the class ``zebra'', the ``has paws'' entry of the attribute vector is zero whereas the ``has stripes'' would be one. The most popular attribute-based prediction algorithm requires learning one classifier per attribute. To classify a new image, its attributes are predicted using the learned classifiers and the attribute scores are combined into class-level scores. This two-step strategy is referred to as Direct Attribute Prediction (DAP) in \cite{LNH09}.

\begin{figure}[t]
  \centering
  \includegraphics[width=\linewidth, trim=0 3cm 0 3.2cm]{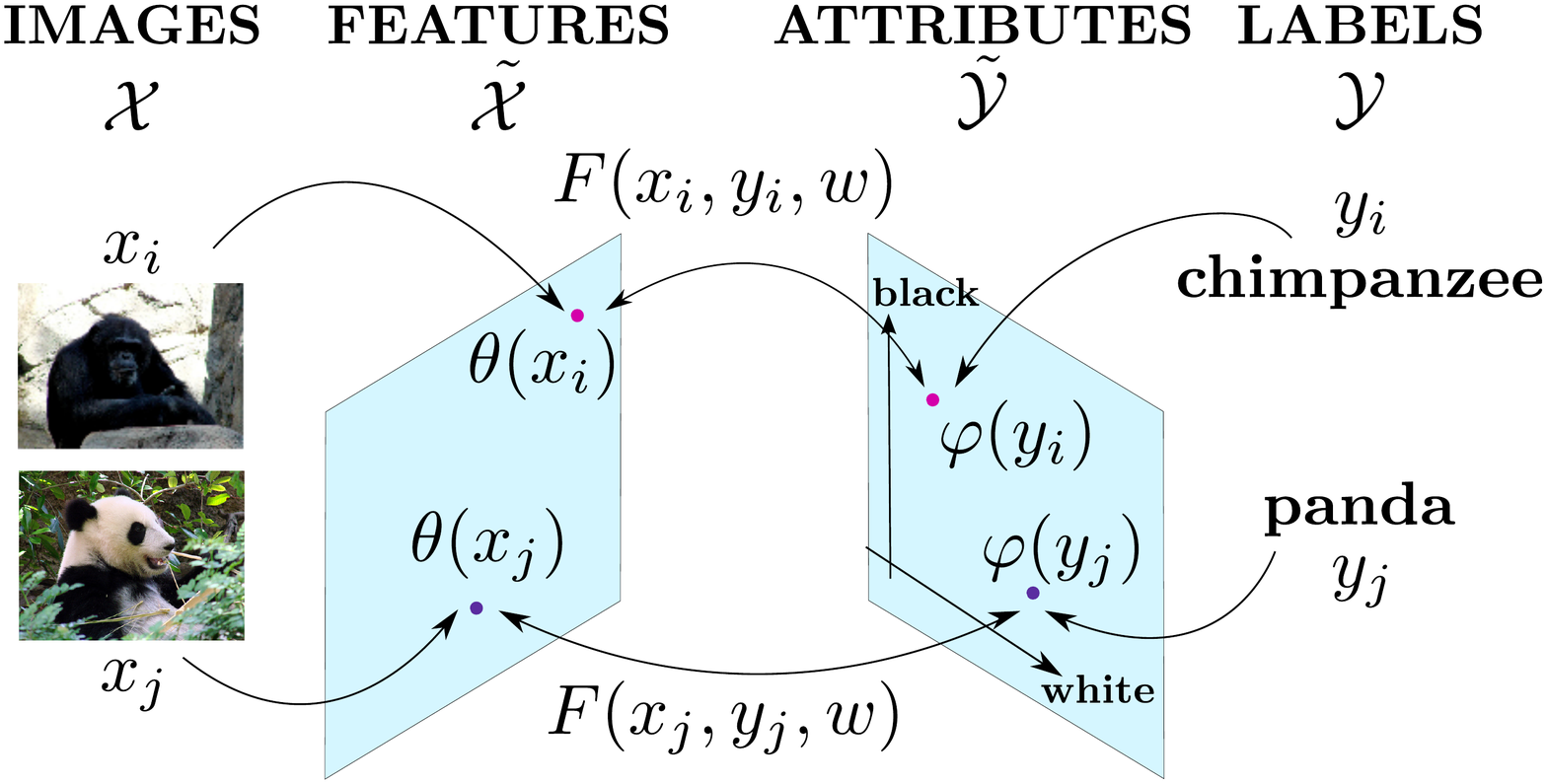}
  \caption{Much work in computer vision has been devoted to image embedding (left): how to extract suitable features from an image. We focus on {\em label embedding} (right): 
how to embed class labels in a Euclidean space. We use side information such as attributes for the label embedding and measure the ``compatibility''' between the embedded inputs and outputs with a function $F$.
}
  \label{fig:ale}
\vspace{-7mm}
\end{figure}

DAP suffers from several shortcomings. First, DAP proceeds in a two-step fashion, learning attribute-specific classifiers in a first step and combining attribute scores into class-level scores in a second step. Since attribute classifiers are learned independently of the end-task the overall strategy of DAP might be optimal at predicting attributes but not necessarily at predicting classes. Second, we would like an approach that can perform zero-shot prediction if no labeled samples are available for some classes, but that can also leverage new labeled samples for these classes as they become available. While DAP is straightforward to implement for zero-shot learning problems, it is not straightforward to extend to such an incremental learning scenario. Third, while attributes can be a useful source of prior information, they are expensive to obtain and the human labeling is not always reliable. Therefore, it is advantageous to seek complementary or alternative sources of side information such as class hierarchies or textual descriptions (see section \ref{sec:bey}). It is not straightforward to design an efficient way to incorporate these additional sources of information into DAP. Various solutions have been proposed to address each of these problems separately (see section \ref{sec:rel}). However, we do not know of any existing solution that addresses all of them in a principled manner.

Our primary contribution is therefore to propose such a solution by making use of the {\em label embedding} framework. We underline that, while there is an abundant literature in the computer vision community on image embedding (how to describe an image) much less work has been devoted in comparison to label embedding in the $\Y$ space (how to describe a class). We embed each class $y \in \Y$ in the space of attribute vectors and thus refer to our approach as \emph{Attribute Label Embedding} (ALE). We use a structured output learning formalism and introduce a function which measures the compatibility between an image $x$ and a label $y$ (see Figure~\ref{fig:ale}). The parameters of this function are learned on a training set of labeled samples to ensure that, given an image, the correct class(es) rank higher than the incorrect ones. Given a test image, recognition consists in searching for the class with the highest compatibility. 

Another important contribution of this work is to show that our approach extends far beyond the setting of attribute-based recognition: it can be readily used for any side information that can be encoded as vectors in order to be leveraged by the label embedding framework. 

Label embedding addresses in a principled fashion the three limitations of DAP that were mentioned previously. First, we optimize directly a class ranking objective, whereas DAP proceeds in two steps by solving intermediate problems. We show experimentally that ALE outperforms DAP in the zero-shot setting. Second, if available, labeled samples can be used to learn the embedding. Third, other sources of side information can be combined with attributes or used as alternative source in place of attributes. 

The paper is organized as follows. In Sec.~2-\ref{sec:le}, we review related work and introduce ALE. In Sec.~\ref{sec:bey}, we study extensions of label embedding beyond attributes. In Sec.~\ref{sec:exp}, we present experimental results on Animals with Attributes (AWA) \cite{LNH09} and Caltech-UCSD-Birds (CUB) \cite{WBPB11}. In particular, we compare ALE with competing alternatives, using the same side information \textit{i.e.} attribute-class associations matrices. 

A preliminary version of this article appeared in \cite{APHS13}. This version adds  (1) an expanded related work section; (2)~ a detailed description of the learning procedure for ALE; (3)~additional comparisons with random embeddings~\cite{DB95} and embeddings derived automatically from textual corpora~\cite{MSCCD13,FCS13}; (4)~additional zero-short learning experiments, which show the advantage of using continuous embeddings; and 
(5)~additional few-shots learning experiments. 

\section{Related work}\label{sec:rel}

We now review related work on attributes, zero-shot learning and label embedding, three research areas which strongly overlap.

\subsection{Attributes}
Attributes have been used for image description \cite{FZ07,FEHF09,CGG12}, caption generation \cite{KPD11,OKB11}, face recognition \cite{KBBN09,SKBB12,CGG13},
image retrieval \cite{KBN08,SFD11,DRS11}, action recognition \cite{LKS11,YJKLGL11},
novelty detection \cite{WB13} and object classification \cite{LNH09,FEHF09,WF09,WM10,MSN11,SQL12,MVC12}. Since our task is object classification in images, we focus on the corresponding references.

The most popular approach to attribute-based recognition is the Direct Attribute Prediction (DAP) model of Lampert \etal which consists in predicting the presence of attributes in an image and combining the attribute prediction probabilities into class prediction probabilities \cite{LNH09}. A significant limitation of DAP is the fact that it assumes that attributes are independent from each other, an assumption which is generally incorrect (see our experiments on attribute correlation in section \ref{sec:zero}).
Consequently, DAP has been improved to take into account the correlation between attributes or between attributes and classes \cite{WF09,WM10,YA10,MSN11}. However, all these models have limitations of their own. Wang and Forsyth \cite{WF09} assume that {\em images} are labeled with both classes and attributes. In our work we only assume that {\em classes} are labeled with attributes, which requires significantly less hand-labeling of the data. Mahajan \etal \cite{MSN11} use transductive learning and, therefore, assume that the test data is available as a batch, a strong assumption we do not make. Yu and Aloimonos's topic model \cite{YA10} is only applicable to bag-of-visual-word image representations and, therefore, cannot leverage recent state-of-the-art image features such as the Fisher vector \cite{SPM13}. We will use such features in our experiments. 
Finally, the latent SVM framework of Wang and Mori \cite{WM10} is not applicable to zero-shot learning, the focus of this work. 

Several works have also considered the problem of discovering a vocabulary of attributes \cite{BBS10,DPCG12,MP13}. \cite{BBS10} leverages text and images sampled from the Internet and uses the mutual information principle to measure the information of a group of attributes. \cite{DPCG12} discovers local attributes and integrates humans in the loop for recommending the selection of attributes that are semantically meaningful. \cite{MP13} discovers attributes from images, textual comments and ratings for the purpose of aesthetic image description. In our work, we assume that the class-attribute association matrix is provided. In this sense, our work is complementary to those previously mentioned.

\subsection{Zero-shot learning}
Zero-shot learning requires the ability to transfer knowledge from classes for which we have training data to classes for which we do not. There are two crucial choices when performing zero-shot learning: the choice of the prior information and the choice of the recognition model.

Possible sources of prior information include attributes \cite{LNH09,FEHF09,PPH09,RSS10,RSS11}, semantic class taxonomies \cite{RSS11,MVP12}, class-to-class similarities \cite{RSS10,YCFSC13}, text features \cite{PPH09,RSS10,RSS11,SGSBMN13,FCS13} or class co-occurrence statistics \cite{MGS14}.
Rohrbach \etal \cite{RSS11} compare different sources of information for learning with zero or few samples. However, since different models are used for the different sources of prior information, it is unclear whether the observed differences are due to the prior information itself or the model. In our work, we compare attributes, class hierarchies and textual information obtained from the internet using the exact same learning framework and  we can, therefore, fairly compare different sources of prior information. Other sources of prior information have been proposed for special purpose problems. For instance, Larochelle \etal \cite{LEB08} encode characters with $7 \times 5$ pixel representations. 
However, it is difficult to extend such an embedding to the case of generic visual categories -- our focus in this work. For a recent survey of different output embeddings optimized for zero-shot learning on fine-grained datasets, the reader may refer to~\cite{ARWLS15}.

As for the recognition model, there are several alternatives. As mentioned earlier, DAP uses a probabilistic model which assumes attribute independence \cite{LNH09}. Closest to the proposed ALE are those works where zero-shot recognition is performed by assigning an image to its closest class embedding (see next section). The measure of distance between an image and a class embedding is generally measured as the Euclidean distance and a transformation is learned to map the input image features to the class embeddings \cite{PPH09,SGSBMN13}. The main difference between these works and ours is that we learn the input-to-output mapping features to optimize directly an image classification criterion: we learn to rank the correct label higher than incorrect ones. We will see in section \ref{sec:zero} that this leads to improved results compared to those works which optimize a regression criterion such as \cite{PPH09,SGSBMN13}.

Few works have considered the problem of transitioning from zero-shot learning to learning with few shots \cite{YA10,SQL12,YCFSC13}. As mentioned earlier, \cite{YA10} is only applicable to bag-of-words type of models. \cite{SQL12} proposes to augment the attribute-based representation with additional dimensions for which an autoencoder model is coupled with a large margin principle. While this extends DAP to learning with labeled data, this approach does not improve DAP for zero-shot recognition. In contrast, we show that the proposed ALE can transition from zero-shot to few-shots learning {\em and} improves on DAP in the zero-shot regime. \cite{YCFSC13} learns separately the class embeddings and the input-to-output mapping which is suboptimal. In this paper,  we learn {\em jointly} the class embeddings (using attributes as prior) and the input-to-output mapping to optimize classification accuracy.

\subsection{Label embedding}
In computer vision, a vast amount of work has been devoted to input embedding, \ie how to represent an image. This includes work on patch encoding (see \cite{CLV11} for a recent comparison), on kernel-based methods \cite{Shawe:Cristianini:2004} with a recent focus on explicit embeddings~\cite{MB09,VZ10}, on dimensionality reduction~\cite{Shawe:Cristianini:2004} and on compression~\cite{JDS11,SP11,VZ12}. Comparatively, much less work has been devoted to label embedding.

Provided that the embedding function $\p$ is chosen correctly -- \ie ``similar'' classes are close according to the Euclidean metric in the embedded space -- label embedding can be an effective way to share parameters between classes. Consequently, the main applications have been multiclass classification with many classes\cite{AFSU07,WC08,WBU10,BWG10} and zero-shot learning \cite{LEB08,PPH09}. We now provide a taxonomy of embeddings. While this taxonomy is valid for both input $\t$ and output embeddings $\p$, we focus here on output embeddings. They can be (i) fixed and data-independent, (ii) learned from data, or (iii) computed from side information. 

\vspace{2mm}\noindent
{\bf Data-Independent Embeddings.}
Kernel dependency estimation~\cite{WestonCESV02} is an example of a strategy where $\p$ is data-independent and defined implicitly through a kernel in the $\Y$ space. The compressed sensing approach of Hsu \etal ~\cite{HKL09}, is another example of data-independent embeddings where $\p$ corresponds to random projections. The Error Correcting Output Codes (ECOC) framework encompasses a large family of embeddings that are built using information-theoretic arguments~\cite{H1950}. ECOC approaches allow in particular to tackle multi-class learning problems as described by Dietterich and Bakiri in \cite{DB95}. 
The reader can refer to \cite{EPR10} for a summary of ECOC methods and latest developments in the ternary output coding methods. Other data-independent embeddings are based on pairwise coupling and variants thereof such as generalized Bradley-Terry models~\cite{Hastie:Tibshirani:Friedman:2008}.

\vspace{2mm}\noindent
{\bf Learned Embeddings.}
A strategy consists in learning jointly $\t$ and $\p$ to embed the inputs and outputs in a common intermediate space ${\cal Z}$. The most popular example is Canonical Correlation Analysis (CCA)~\cite{Hastie:Tibshirani:Friedman:2008}, which maximizes the correlation between inputs and outputs. Other strategies have been investigated which maximize directly classification accuracy, including the nuclear norm regularized learning of Amit \etal~\cite{AFSU07} or the WSABIE algorithm of Weston \etal~\cite{WBU10}.

\vspace{2mm}\noindent
{\bf Embeddings Derived From Side Information.}
There are situations where side information is available. This setting is particularly relevant when little training data is available, as side information and the derived embeddings can compensate for the lack of data. Side information can be obtained at an image level \cite{FEHF09} or at a class level \cite{LNH09}. We focus on the latter setting which is more practical as collecting side information at an image level is more costly. Side information may include ``hand-drawn'' descriptions \cite{LEB08}, text descriptions \cite{FEHF09,LNH09,PPH09,FCS13} or class taxonomies \cite{WC08,BWG10}. Certainly, the closest work to ours is that of Frome \etal~\cite{FCS13}~\footnote{Note that the work of Frome \etal~\cite{FCS13} is posterior to our conference submission~\cite{APHS13}.} which involves embedding classes using textual corpora and then learning a mapping between the input and output embeddings using a ranking objective function. We also use a ranking objective function and compare different sources of side information to perform embedding: attributes, class taxonomies and textual corpora.

\vspace{2mm}

While our focus is on embeddings derived from side information for zero-shot recognition,
we also considered data independent embeddings and learned embeddings (using side information as a prior) for few-shots recognition.

\section{Label embedding with attributes}
\label{sec:le}

Given a training set ${\cal S}=\{(x_n,y_n), n =1 \ldots N\}$ of input/output pairs with $x_n \in \X$ and $y_n \in \Y$, our goal is to learn a function $f: \X \rightarrow \Y$ by minimizing an empirical risk of the form
\begin{equation}
\min_{f \in \mathcal{F}}\quad \frac{1}{N} \sum_{n=1}^N \D(y_n,f(x_n))
\end{equation}
where $\D: \Y \times \Y \rightarrow \Re$ measures the loss incurred from predicting $f(x)$ when the true label is $y$, and where the function $f$ belongs to the function $\mathcal{F}$. We shall use the 0/1 loss as a target loss: $\D(y,z) = 0$ if $y=z$, 1 otherwise, to measure the test error, while we consider several surrogate losses commonly used for structured prediction at learning time (see Sec.~\ref{sec:obj} for details on the surrogate losses used in this paper).

An elegant framework, initially proposed in~\cite{WestonCESV02}, allows to concisely describe learning problems where both input and output spaces are jointly or independently mapped into lower-dimensional spaces. The framework relies on so-called \emph{embedding functions} $\theta: \X \rightarrow \tX$ and $\p: \Y \rightarrow \tY$ resp for the inputs and outputs. Thanks to these embedding functions, the learning problem is cast into a regular learning problem with transformed input/output pairs. 

In what follows, we first describe our function class $\mathcal{F}$ (section\ref{sec:frm}). We then explain how to leverage side information under the form attributes to compute label embeddings (section \ref{sec:ale}). We also discuss how to learn the model parameters (section \ref{sec:obj}). While, for the sake of simplicity, we focus on attributes in this section, the approach readily generalizes to any side information that can be encoded in matrix form (see following section \ref{sec:bey}).

\subsection{Framework}
\label{sec:frm}
Figure~\ref{fig:ale} illustrates the proposed model. Inspired from the structured prediction formulation \cite{TJH05}, we introduce a compatibility function $F: \X \times \Y \rightarrow \Re$ and define $f$ as follows:
\begin{equation}
f(x;w) = \arg \max_{y \in \Y} F(x,y; w)
\label{eq:annot}
\end{equation}
where $w$ denotes the model parameter vector of $F$ and $F(x,y;w)$ measures how compatible is the pair $(x,y)$ given $w$. It is generally assumed that $F$ is linear in some combined feature embedding of inputs/outputs $\psi(x,y)$:
\begin{equation}
F(x,y; w) = w'\psi(x,y)
\end{equation}
and that the joint embedding $\psi$ can be written as the tensor product between the image embedding $\theta: \X \rightarrow \tX = \Re^D$ and the label embedding $\p: \Y \rightarrow \tY = \Re^E$:
\begin{equation}
\psi(x,y) = \theta(x) \otimes \p(y) 
\end{equation}
and $\psi(x,y): \Re^D \times \Re^E \rightarrow \Re^{D E}$. In this case $w$ is a DE-dimensional vector which can be reshaped into a $D \times E$ matrix $W$. Consequently, we can rewrite $F(x,y;w)$ as a bilinear form:
\begin{equation}
F(x,y;W) = \theta(x)' W \p(y) . 
\label{eqn:form}
\end{equation}
Other compatibility functions could have been considered. For example, the function:
\begin{equation}
F(x,y;W) = - \Vert \theta(x)'W - \p(y) \Vert^2
\label{eq:reg}
\end{equation}
is typically used in regression problems. 

Also, if $D$ and $E$ are large, it might be valuable to consider a low-rank decomposition $W=U'V$ to reduce the effective number of parameters. In such a case, we have:
\begin{equation}
F(x,y;U,V) = \left( U \theta(x)\right)' \left( V \p(y) \right). 
\label{eqn:lowrank}
\end{equation}
CCA~\cite{Hastie:Tibshirani:Friedman:2008}, or more recently WSABIE~\cite{WBU10} rely, for example, on such a decomposition. 

\subsection{Embedding classes with attributes}
\label{sec:ale}
We now consider the problem of defining the label embedding function $\p^{\cal A}$ from attribute side information. In this case, we refer to our approach as \textbf{Attribute Label Embedding (ALE)}.

We assume that we have $C$ classes, \ie $\Y = \{1, \ldots, C\}$ and that we have a set of $E$ attributes ${\cal } = \{a_i, i=1 \ldots E\}$ to describe the classes. We also assume that we are provided with an association measure $\rho_{y,i}$ between each attribute $a_i$
and each class $y$. These associations may be binary or real-valued if we have information
about the association strength (\eg if the association value is obtained by averaging votes). We embed class $y$ in the $E$-dim attribute space as follows:
\begin{equation}
\p^{\cal A}(y) = [\rho_{y,1}, \ldots, \rho_{y,E}]
\end{equation}
and denote $\Phi^{\cal A}$ the $E \times C$ matrix of attribute embeddings which stacks the individual $\p^{\cal A}(y)$'s.

We note that in equation (\ref{eqn:form}) the image and label embeddings play symmetric roles. In the same way it makes sense to normalize samples when they are used as input to large-margin classifiers, it can make sense to normalize the output vectors $\p^{\cal A}(y)$. In section \ref{sec:zero} we compare (i) continuous embeddings, (ii) binary embeddings using $\{0,1\}$ for the encoding and (iii) binary embeddings using $\{-1,+1\}$ for the encoding. We also explore two normalization strategies: (i) mean-centering (\ie compute the mean over all learning classes and subtract it) and (ii) $\ell_2$-normalization. We underline that such encoding and normalization choices are not arbitrary but relate to prior assumptions we might have on the problem. For instance, underlying the $\{0,1\}$ embedding is the assumption that the presence of the same attribute in two classes should contribute to their similarity, but not its absence. Here we assume a dot-product similarity between attribute embeddings which is consistent with our linear compatibility function (\ref{eqn:form}). Underlying the $\{-1,1\}$ embedding is the assumption that the presence or the absence of the same attribute in two classes should contribute equally to their similarity. As for mean-centered attributes, they take into account the fact that some attributes are more frequent than others. For instance, if an attribute appears in almost all classes, then in the mean-centered embedding, its absence will contribute more to the similarity than its presence. This is similar to an IDF effect in TF-IDF encoding. As for the $\ell_2$-normalization, it enforces that each class is closest to itself according to the dot-product similarity.

In the case where attributes are redundant, it might be advantageous to de-correlate them. In such a case, we make use of the compatibility function (\ref{eqn:lowrank}). The matrix $V$ may be learned from labeled data jointly with $U$. As a simpler alternative, it is possible to first learn the decorrelation, \eg by performing a Singular Value Decomposition (SVD) on the $\Phi^{\cal A}$ matrix, and then to learn $U$. We will study the effect of attribute de-correlation in our experiments.

\subsection{Learning algorithm}
\label{sec:obj}

We now turn to the estimation of the model parameters $W$ from a labeled training set ${\cal S}$. The simplest learning strategy is to maximize directly the compatibility between the input and output embeddings:
\begin{equation}
 \frac{1}{N} \sum_{n=1}^N F(x_n,y_n;W)
\end{equation}
with potentially some constraints and regularizations on $W$. This is exactly the strategy adopted in regression \cite{PPH09,SGSBMN13}. However, such an objective function does not optimize directly our end-goal which is image classification. Therefore, we draw inspiration from the WSABIE algorithm \cite{WBU10} that learns jointly image and label embeddings from data to optimize classification accuracy.
The {\em crucial difference between WSABIE and ALE is the fact that the latter uses attributes as side information}. Note that {\em the proposed ALE is not tied to WSABIE} and that we report results in \ref{sec:zero} with other objective functions including regression and structured SVM (SSVM). We chose to focus on the WSABIE objective function with ALE because it yields good results and is scalable.

In what follows, we briefly review the WSABIE objective function~\cite{WBU10}. Then, we present ALE which allows to do (i) zero-shot learning with side information and (ii) learning with few (or more) examples with side information. We, then, detail the proposed learning procedures for ALE. In what follows, $\Phi$ is the matrix which stacks the embeddings $\p(y)$.

\vspace{2mm}\noindent
{\bf WSABIE.}
Let $\1(u)$ = 1 if $u$ is true and 0 otherwise. Let:
\begin{equation}
\ell(x_n,y_n,y) = \D(y_n,y) + \theta(x)'W[\p(y) -\p(y_n)]
\end{equation}
Let $r(x_n,y_n)$ be the rank of label $y_n$ for image $x_n$. Finally, let $\a_1, \a_2, \ldots, \a_C$ be a sequence of $C$ non-negative coefficients and let $\b_k = \sum_{j=1}^k \a_j$. Usunier \etal \cite{UBG09} propose to use the following ranking loss for ${\cal S}$:
\begin{equation} \label{eqn:owa}
\frac{1}{N} \sum_{n=1}^N \b_{r(x_n,y_n)} \; ,
\end{equation}
where $\b_{r(x_n,y_n)} := \sum_{j=1}^{r(x_n,y_n)} \alpha_j$. Since the $\b_k$'s are increasing with $k$, minimizing $\b_{r(x_n,y_n)}$ enforces to minimize the $r(x_n,y_n)$'s, \ie it enforces correct labels to rank higher than incorrect ones. $\a_k$ quantifies the penalty incurred by going from rank $k$ to $k+1$. Hence, a decreasing sequence $\a_1 \geq \a_2 \geq \ldots \geq \a_C \geq 0$ implies that a mistake on the rank when the true rank is at the top of the list incurs a higher loss than a mistake on the rank when the true rank is lower in the list -- a desirable property. Following Usunier \etal, we choose $\a_k = 1/k$.

Instead of optimizing an upper-bound on (\ref{eqn:owa}), Weston \etal propose to optimize the following approximation of objective (\ref{eqn:owa}):
\begin{equation}
R({\cal S};W,\Phi) = \frac{1}{N} \sum_{n=1}^N \frac{\b_{r_\D(x_n,y_n)}}{r_{\D(x_n,y_n)}} \sum_{y \in \Y} \max\{0,\ell(x_n,y_n,y)\}
\label{eqn:wsabie}
\end{equation}
where
\begin{equation}
r_{\D}(x_n,y_n) =  \sum_{y \in \Y} \1(\ell(x_n,y_n,y)>0) 
\end{equation}
is an upper-bound on the rank of label $y_n$ for image $x_n$.

The main advantage of the formulation (\ref{eqn:wsabie}) is that it can be optimized efficiently through Stochastic Gradient Descent (SGD), as described in Algorithm \ref{alg:sgd}. The label embedding space dimensionality is a parameter to set, for instance using cross-validation. Note that the previous objective function does not incorporate any regularization term. Regularization is achieved implicitly by early stopping, \ie the learning is terminated once the accuracy stops increasing on the validation set.

\vspace{2mm}
\noindent {\bf ALE: Zero-Shot Learning.}
We now describe the ALE objective for zero-shot learning. In such a case, we cannot learn $\Phi$ from labeled data, but rely on side information. This is in contrast to WSABIE. 
Therefore, the matrix $\Phi$ is fixed and set to $\Phi^{\cal A}$ (see section \ref{sec:ale} for details on $\Phi^{\cal A}$). We only optimize the objective (\ref{eqn:wsabie}) with respect to $W$.
We note that, when $\Phi$ is fixed and only $W$ is learned, the objective (\ref{eqn:wsabie}) is closely related to the (unregularized) structured 
SVM (SSVM) objective \cite{TJH05}:
\begin{equation}
\frac{1}{N} \sum_{n=1}^N \max_{y \in \Y} \ell(x_n,y_n,y)
\label{eqn:ssvm}
\end{equation}
The main difference is the loss function, which is the multi-class loss function for SSVM. The multi-class loss function focuses on the score with the highest rank, while ALE considers all scores in a weighted fashion. Similar to WSABIE, a major advantage of ALE is its scalability to large datasets \cite{WBU10,PAH12}.

\vspace{2mm}
\noindent {\bf ALE: Few-Shots Learning.}
We now describe the ALE objective to the case where we have labeled data and side information. In such a case, we want to learn the class embeddings using as prior information $\Phi^{\cal A}$. We, therefore, add to the objective (\ref{eqn:wsabie}) a regularizer:
\begin{equation}
R({\cal S};W,\Phi) +\frac{\mu}{2} ||\Phi - \Phi^{\cal A}||^2
\label{eqn:ale_fewshots}
\end{equation}
and optimize jointly with respect to $W$ and $\Phi$. Note that the previous equation is somewhat reminiscent of the ranking model adaptation of \cite{GYX12}.

\vspace{2mm}\noindent
{\bf Training.}
For the optimization of the zero-shot as well as the few-shots learning, we follow \cite{WBU10} and use Stochastic Gradient Descent (SGD). Training with SGD consists at each step $t$ in (i) choosing a sample $(x,y)$ at random, (ii) repeatedly sampling a negative class denoted $\bar{y}$ with $\bar{y} \neq y$ until a violating class is found, \ie until $\ell(x,y,\bar{y}) > 0$, and (iii) updating the projection matrix (and the class embeddings in case of few-shots learning) using a sample-wise estimate of the regularized risk. Following \cite{WBU10,PAH12}, we use a constant step size $\eta_t=\eta$.
The detailed algorithm is provided in Algorithm \ref{alg:sgd}.
\begin{algorithm}[t]
	\small
  \caption{ALE stochastic training}
  Intitialize $W^{(0)}$ randomly.
  \begin{algorithmic}
    \For  {$t=1$ to $T$} 
      \State  Draw ($x$,$y$) from ${\cal S}$. 
      \For  {$k=1, 2, \ldots, C-1$} 
        \State  Draw $\bar{y} \neq y$ from  $\Y$  
        \If {$\ell(x,y,\bar{y}) > 0$} 
          \State {\bf // Update $W$}
          \begin{equation} W^{(t)} = W^{(t-1)} + \eta_t \beta_{\lfloor \frac{C-1}{k} \rfloor} \theta(x) [\p(y) - \p(\bar{y})]' \end{equation}
          \State {\bf // Update $\Phi$ (not applicable to zero-shot)}
          \begin{eqnarray} \p^{(t)}(y) & = & (1 - \eta_t \mu) \p^{(t-1)}(y)  + \eta_t \mu \p^{{\cal A}}(y) \nonumber \\ 
            & + & \eta_t \beta_{\lfloor \frac{C-1}{k} \rfloor} W' \theta(x) \label{eqn:upd1} \end{eqnarray}
          \begin{eqnarray} \p^{(t)}(\bar{y}) & = & (1 - \eta_t \mu) \p^{(t-1)}(\bar{y}) + \eta_t \mu \p^{{\cal A}}(\bar{y}) \nonumber \\
            & - & \eta_t \beta_{\lfloor \frac{C-1}{k} \rfloor} W' \theta(x) \label{eqn:upd2} \end{eqnarray}
        \EndIf
      \EndFor
    \EndFor
  \end{algorithmic}
  \label{alg:sgd}
\end{algorithm} 

\section{Label embedding beyond attributes}
\label{sec:bey}

A wealth of label embedding methods have been proposed over the years, in several communities and most often for different purpose. Previous works considered either fixed (data-independent) or learned-from-data embeddings. Data used for learning could be either \emph{restricted to the task-at-hand} or could also be complemented by \emph{side information} from other modalities. The purpose of this paper is to propose a general framework that encompasses all these approaches, and compare the empirical performance on image classification tasks. Label embedding methods could be organized according to two criteria: i) task-focused or using other sources of side information; ii) fixed or data-dependent embedding. 

\subsection{Side information in label embedding}
A first criterion to discriminate among the different approaches for label embedding is whether the method is using only the training data for the task at hand, that is the examples (images) along with their class labels, or if it is using other sources of information. In the latter option, side information impacts the outputs, and can rely on several types of modalities. In our setting, these modalities could be i) attributes, ii) class taxonomies or iii) textual corpora. i) was the focus of the previous section (see especially \ref{sec:ale}). In what follows, we focus on ii) and iii).

\begin{figure}[t]
  \centering
  \includegraphics[width=0.42\linewidth]{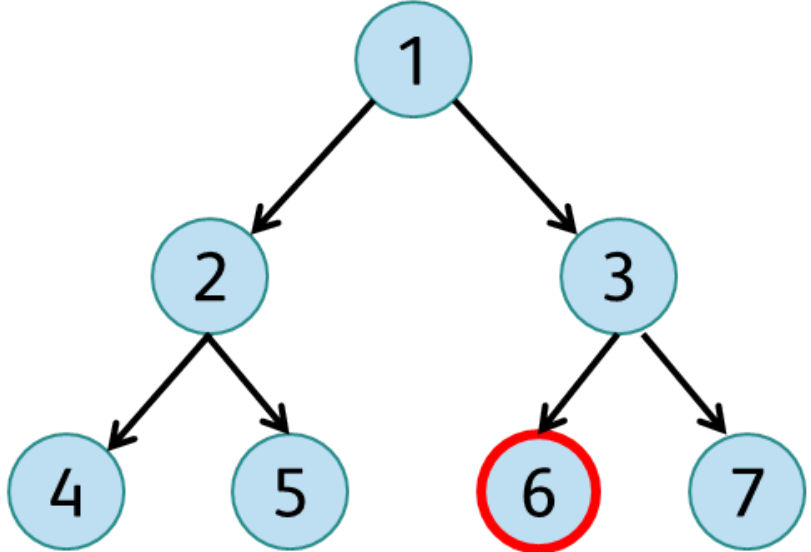}
  \caption{Illustration of Hierarchical Label Embedding (HLE).
In this example, given 7 classes (including a ``root'' class),
class $6$ is encoded in a binary 7-dimensional space as 
$\p^{\cal H}(6) = [1, 0, 1, 0, 0, 1, 0]$.}
  \label{fig:hie}
\vspace{-4mm}
\end{figure}

\vspace{2mm}\noindent
{\bf Class hierarchical structures} explicitly use expert knowledge to group the image classes into a hierarchy, such as knowledge from ornithology for birds datasets. A hierarchical structure on the classes requires an ordering operation $\prec$ in $\Y$: $z \prec y$ means that $z$ is an ancestor of $y$ in the tree hierarchy. Given this tree structure, we can define $\xi_{y,z} = 1$ if $z \prec y$ or $z = y$. The hierarchy embedding $\p^{\cal H}(y)$ can be defined as the $C$ dimensional vector:
\begin{equation}
\p^{\cal H}(y) = [\xi_{y,1}, \ldots, \xi_{y,C}] .
\end{equation}
Here, $\xi_{y,i}$ is the association measure of the $i^{th}$ node in the hierarchy with class $y$. See Figure~\ref{fig:hie} for an illustration. We refer to this embedding as \textbf{Hierarchy Label Embedding (HLE)}. Note that HLE was first proposed in the context of structured learning \cite{TJH05}. Note also that, if classes are not organized in a tree structure but form a graph, other types of embeddings can be used, for instance by performing a kernel PCA on the commute time kernel \cite{SFY04}.

\vspace{2mm}\noindent
{\bf The co-occurrence of class names in textual corpora} can be automatically extracted using field guides or public resources such as Wikipedia~\footnote{\url{http://en.wikipedia.org}}. Co-occurences of class names can be leveraged to infer relationships between classes, leading to an embedding of the classes.  Standard approaches to produce word embeddings from co-ocurrences include Latent Semantic Analyis (LSA)~\cite{De88}, probabilistic Latent Semantic Analysis (pLSA)~\cite{Ho99} or Latent Dirichlet Allocation (LDA)~\cite{BNJ03}. In this work, we use the recent state-of-the-art approach of Mikolov \etal ~\cite{MSCCD13}, also referred to as ``Word2Vec''. It uses a skip-gram model that enforces a word (or a phrase) to  be a good predictor of its surrounding words, \ie it enforces neighboring words (or phrases) to be close to each other in the embedded space. Such an embedding , which we refer to as \textbf{Word2Vec Label Embedding (WLE)}, was recently used for zero-shot recognition~\cite{FCS13} on fine-grained datasets~\cite{ARWLS15}.

\vspace{2mm}
In section \ref{sec:exp}, we compare attributes, class hierarchies and textual information (\ie resp. ALE, HLE and WLE) as sources of side information for zero-shot recognition.

\subsection{Data-dependence of label embedding}
A second criterion is whether the label embedding used at prediction time was fit to training data at training time or not. Here, being \emph{data-dependent} refers to the \emph{training data}, putting aside all other possibles sources of information. There are several types of approaches in this respect: i) fixed and data-independent label embeddings; ii) data-dependent, learnt solely from training data; iii) data-dependent, learnt jointly from training data and side information. 

Fixed and data-independent correspond to fixed mappings of the original class labels to a lower-dimensional space. In our experiments, we explore three of such kind of embeddings: i) trivial label embedding corresponding to identity mapping, which boils down to plain one-versus-rest classification (\textbf{OVR}); ii) Gaussian Label Embedding (\textbf{GLE}), using Gaussian random projection matrices and assuming Johnson-Lindenstrauss properties; iii) Hadamard Label embedding, similarly, using Hadamard matrices for building the random projection matrices. None of these three label embedding approaches use the training data (nor any side information) to build the label embedding. It is worthwhile to note that the underlying dimensions of these label embedding do rely on training data, since they are usually cross-validated; we shall however ignore this fact here for simplicity of the exposition. 

Data-dependent label embedding use the training data to build the label embedding used at prediction time. Popular methods in this family are principal component analysis on the outputs, and canonical correlation analysis, and the plain \textbf{WSABIE} approach. 

Note that it is possible to use both the available training data {\em and} side information to learn the embedding functions. The proposed family of approaches, Attribute Label Embedding (\textbf{ALE}), 
belongs to this latter category.

\vspace{2mm}\noindent
{\bf Combining embeddings.} Different embeddings can be easily combined in the label embedding framework, \eg through simple concatenation of the different embeddings or through more complex operations such as a CCA of the embeddings. This is to be contrasted with DAP which cannot accommodate so easily other sources of prior information.

\section{Experiments}
\label{sec:exp}

We now evaluate the proposed ALE framework on two public benchmarks: Animal With Attributes (AWA) and CUB-200-2011 (CUB). AWA \cite{LNH09} contains roughly 30,000 images of 50 animal classes. CUB \cite{WBPB11} contains roughly 11,800 images of 200 bird classes. 

We first describe in sections~\ref{sec:in} and \ref{sec:out} respectively the input embeddings (\ie image features) and output embeddings that we have used in our experiments. In section~\ref{sec:zero}, we present zero-shot recognition experiments, where training and test classes are disjoint. In section~\ref{sec:few}, we go beyond zero-shot learning and consider the case where we have plenty of training data for some classes and little training data for others. Finally, in section~\ref{sec:full} we report results in the case where we have equal amounts of training data for all classes.

\subsection{Input embeddings}
\label{sec:in}
Images are resized to 100K pixels if larger while keeping the aspect ratio. We extract 128-dim SIFT descriptors \cite{Lo04} and 96-dim color descriptors \cite{CCP07} from regular grids at multiple scales. Both of them are reduced to 64-dim using PCA. These descriptors are, then, aggregated into an image-level representation using the Fisher Vector (FV) \cite{PSM10}, shown to be a state-of-the-art patch encoding technique in~\cite{CLV11}. Therefore, our input embedding function $\theta$ takes as input an image and outputs a FV representation. Using Gaussian Mixture Models with 16 or 256 Gaussians, we compute one SIFT FV and one color FV per image and concatenate them into either 4,096 (4K) or 65,536-dim (64K) FVs. As opposed to \cite{APHS13}, we do not apply PQ-compression which explains why we report better results in the current work (\eg on average 2\% better with the same output embeddings on CUB).

\subsection{Output Embeddings}
\label{sec:out}

In our experiments, we considered three embeddings derived side information: attributes, class taxonomies and textual corpora. When considering attributes, we use the attributes (binary, or continuous) as they are provided with the datasets, with no further side information.

\vspace{2mm}
\noindent {\bf Attribute Label Embedding (ALE).} In AWA, each class was annotated with 85 attributes by 10 students \cite{OSW91}. Continuous class-attribute associations were obtained by averaging the per-student votes and subsequently thresholded to obtain binary attributes. In CUB, 312 attributes were obtained from a bird field guide. Each image was annotated according to the presence/absence of these attributes. The per-image attributes were averaged to obtain continuous-valued class-attribute associations and thresholded with respect to the overall mean to obtain binary attributes. By default, we use continuous attribute embeddings in our experiments on both datasets.

\vspace{2mm}
\noindent {\bf Hierarchical Label Embedding (HLE).} We use the Wordnet hierarchy as a source of prior information to compute output embeddings. We collect the set of ancestors of the 50 AWA (resp. 200 CUB) classes from Wordnet and build a hierarchy with 150 (resp. 299) nodes\footnote{In some cases, some of the nodes have a single child. We did not clean the automatically obtained hierarchy.}. Hence, the output dimensionality is 150 (resp. 299) for AWA (resp. CUB). We compute the binary output codes following~\cite{TJH05}: for a given class, an output dimension is set to $\{0,1\}$ according the absence/presence of the corresponding node among the ancestors. The class embeddings are subsequently $\ell_2$-normalized.

\vspace{2mm}
\noindent {\bf Word2Vec Label Embedding (WLE).}
We trained the skip-gram model on the 13 February 2014 version of the English-language Wikipedia which was tokenized to 1.5 million words and phrases that contain the names of our visual object classes. 
Additionally we use a hierarchical softmax layer \footnote{We obtain word2vec representations using the publicly available implementation from \texttt{https://code.google.com/p/word2vec/}.}. The dimensionality of the output embeddings was cross-validated on a per-dataset basis.

\vspace{2mm}
We also considered three data-independent embeddings:

\vspace{2mm}
\noindent {\bf One-Vs-Rest embedding (OVR).}
The embedding dimensionality is $C$ where $C$ is the number of classes and the matrix $\Phi$ is the $C \times C$ identity matrix. This is equivalent to training independently one classifier per class.

\vspace{2mm}
\noindent {\bf Gaussian Label Embedding (GLE).}
The class embeddings are drawn from a standard normal distribution, similar to random projections in compressed sensing~\cite{DeVore:2007}. Similarly to WSABIE, the label embedding dimensionality $E$ is a parameter of GLE which needs to be cross-validated. For GLE, since the embedding is randomly drawn, we repeat the experiments 10 times and report the average (as well as the standard deviation when relevant).

\vspace{2mm}
\noindent {\bf Hadamard Label Embedding.}
An Hadamard matrix is a square matrix whose rows/columns are mutually orthogonal and whose entries are $\{-1,1\}$~\cite{DeVore:2007}. Hadamard matrices can be computed iteratively with $H_1 = (1)$ and $H_{2^k} = \left( \begin{array}{cc} H_{2^{k-1}} & H_{2^{k-1}}\\ 
H_{2^{k-1}} & -H_{2^{k-1}} \end{array} \right)$. In our experiments Hadamard embedding yielded significantly worse results than GLE. Therefore, we only report GLE results in the following. 

\vspace{2mm}
Finally, when labeled training data is available in sufficient quantity, the embeddings can be learned from the training data. In this work, we considered one data-driven approach to label embedding:

\vspace{2mm}
\noindent {\bf Web-Scale Annotation By Image Embedding (WSABIE).}
The objective function of WSABIE~\cite{WBU10} is provided in (\ref{eqn:wsabie}) and the corresponding optimization algorithm is similar to the one of ALE described in Algorithm 1. The difference is that WSABIE does not use any prior information and, therefore, the regularization value $\mu$ is set to 0 in equations (\ref{eqn:upd1}) and (\ref{eqn:upd2}). Another difference with ALE is that the embedding dimensionality $E$ is a parameter of WSABIE which is obtained through cross-validation. This is an advantage of WSABIE since it provides an additional free parameter compared to ALE. However, the cross-validation procedure is computationally intensive. 

\vspace{2mm}
In summary, in the following we report results for six label embedding strategies: ALE, HLE, WLE, OVR, GLE and WSABIE. Note that OVR, GLE and WSABIE are not applicable to zero-shot learning since they do not rely on any source of prior information and consequently do not provide a meaningful way to embed a new class for which we do not have any training data.

\subsection{Zero-Shot Learning}
\label{sec:zero}

\vspace{2mm} \noindent
{\bf Set-up.} In this section, we evaluate the proposed ALE in the zero-shot setting. For AWA, we use the standard zero-shot setup which consists in learning parameters on 40 classes and evaluating accuracy on the 10 remaining ones. We use all the images in 40 learning classes ($\approx$ 24,700 images) to learn and cross-validate the model parameters. We then use all the images in 10 evaluation classes ($\approx$ 6,200 images) to measure accuracy. For CUB, we use 150 classes for learning ($\approx$ 8,900 images) and 50 for evaluation ($\approx$ 2,900 images). 

\begin{table}[t]
 \begin{center}
  \small
  \resizebox{\linewidth}{!}{
  \begin{tabular}{|l|l||c|c|c|c|c|c|}
		\hline
		& & \multicolumn{6}{c|}{AWA} \\
		\hline
		& & \multicolumn{3}{c|}{FV=4K} & \multicolumn{3}{c|}{FV=64K}\\
		\hline
		$\mu$ & $\ell_2$ & cont & $\{0,1\}$ & $\{-1,+1\}$  & cont & $\{0,1\}$ & $\{-1,+1\}$ \\
		\hline
		no & no &  41.5 & 34.2 & 32.5 &  44.9 & 42.4 & 41.8 \\
		yes & no & 42.2 & 33.8 & 33.8 & 44.9 & 42.4 & 42.4 \\
		no & yes & {\bf 45.7} & 34.2 & 34.8 & {\bf 48.5} & 44.6 & 41.8 \\
		yes & yes & 44.2 & 34.9 & 34.9 & 47.7 & 44.8 & 44.8 \\
		\hline
		\hline
		& & \multicolumn{6}{c|}{CUB} \\
		\hline
		& & \multicolumn{3}{c|}{FV=4K} & \multicolumn{3}{c|}{FV=64K}\\
		\hline
		$\mu$ & $\ell_2$ & cont & $\{0,1\}$ & $\{-1,+1\}$  & cont & $\{0,1\}$ & $\{-1,+1\}$ \\
		\hline
		no & no &  17.2 & 10.4 & 12.8 &  22.7 & 20.5 & 19.6 \\
		yes & no & 16.4 & 10.4 & 10.4 & 21.8 & 20.5 & 20.5 \\
		no & yes & {\bf 20.7} & 15.4 & 15.2 & {\bf 26.9} & 22.3 & 19.6 \\
		yes & yes & 20.0 & 15.6 & 15.6 & 26.3 & 22.8 & 22.8 \\
		\hline
   \end{tabular}
   }
  \end{center}

\caption{Comparison of the continuous embedding (cont), the binary $\{0,1\}$ embedding and the binary $\{+1,-1\}$ embedding. We also study the impact of mean-centering ($\mu$) and $\ell_2$-normalization.}
\label{tab:enc} \vspace{-7mm}
\end{table}

\vspace{2mm}\noindent
{\bf Comparison of output encodings for ALE.} We first compare three different output encodings: (i) continuous encoding, \ie we do not binarize the class-attribute associations, (ii) binary $\{0,1\}$ encoding and (iii) binary $\{-1,+1\}$ encoding. We also compare two normalizations: (i) mean-centering of the output embeddings and (ii) $\ell_2$-normalization. We use the same embedding and normalization strategies at training and test time.

Results are shown in Table \ref{tab:enc}. The conclusions are the following ones. Significantly better results are obtained with continuous embeddings than with thresholded binary embeddings. On AWA with 64K-dim FV, the accuracy is 48.5\% with continuous and 41.8\% with $\{-1,+1\}$ embeddings. Similarly on CUB with 64K-dim FV, we obtain 26.9\% with continuous and 19.6\% with $\{-1,+1\}$ embeddings. This is expected since continuous embeddings encode the strength of association between a class and an attribute and, therefore, carry more information. We believe that this is a major strength of the proposed approach as other algorithms such as DAP cannot accommodate such soft values in a straightforward manner. Mean-centering seems to have little impact with 0.8\% (between 48.5\% and 47.7\%) on AWA and 0.6\% (between 26.9\% and 26.3\%) on CUB using 64K FV as input and continuous attributes as output embeddings.  On the other hand, $\ell_2$-normalization makes a significant difference in all configurations except from the $\{-1,+1\}$ encoding (\eg only 2.4\% difference between 44.8\% and 42.4\% on AWA, 2.3\% difference between 22.8\% and 20.5\% on CUB). This is expected, since all class embeddings already have a constant norm for $\{-1,+1\}$ embeddings (the square-root of the number of output dimensions $E$). In what follows, we always use the continuous $\ell_2$-normalized embeddings without mean-centric normalization.

\begin{table}[t]
 \begin{center}
  \small
  \begin{tabular}{|r||c|c|c|}
\hline
& RR & SSVM & RNK \\
\hline
AWA & 44.5  & 47.9 & \bf{48.5} \\
\hline
CUB & 21.6  & {\bf 26.3} & \bf{26.3} \\
\hline
\end{tabular}
\end{center}
\caption{Comparison of different learning algorithms for ALE: ridge-regression (RR), multi-class SSVM (SSVM) and ranking based on WSABIE (RNK).}
\label{tab:learn} \vspace{-5mm}
\end{table}

\vspace{2mm}\noindent
{\bf Comparison of learning algorithms.} We now compare three objective functions to learn the mapping between inputs and outputs. The first one is Ridge Regression (RR) which was used in \cite{PPH09} to map input features to output attribute labels. In a nutshell, RR consists in optimizing a regularized quadratic loss for which there exists a closed form formula. The second one is the standard structured SVM (SSVM) multiclass objective function of \cite{TJH05}. The third one is the ranking objective (RNK) of WSABIE \cite{WBU10} which is described in detail section \ref{sec:obj}. The results are provided in Table \ref{tab:learn}. On AWA, the highest result is 48.5\% obtained with RNK, followed by MUL with 47.9\% whereas RR performs worse with 44.5\%. On CUB, RNK and MUL obtain 26.3\% accuracy whereas RR again performs somewhat worse with 21.6\%. Therefore, the conclusion is that the multiclass and ranking frameworks are on-par and outperform the simple ridge regression. This is not surprising since the two former objective functions are more closely related to our end goal which is classification. In what follows, we always use the ranking framework (RNK) to learn the parameters of our model, since it both performs well and was shown to be scalable \cite{WBU10,PAH12}.

\begin{table}[t]
 \begin{center}
  \small
  \begin{tabular}{|r|c|c||c|c|}
\hline
& \multicolumn{2}{|c||}{Obj. pred.} & \multicolumn{2}{|c|}{Att. pred.} \\
\hline
& DAP & ALE & DAP & ALE \\
\hline
\hline
AWA & 41.0 & \bf{48.5} & \bf{72.7} & \bf{72.7} \\
\hline
CUB & 12.3 & \bf{26.9} & \bf{64.8} & 59.4 \\
\hline
\end{tabular}
\end{center}
\caption{Comparison of DAP \cite{LNH09} with ALE. Left: object classification accuracy (top-1 \%) on the 10 AWA and 50 CUB evaluation classes. Right: attribute prediction accuracy (AUC \%) on the 85 AWA and 312 CUB attributes. We use 64K FVs.}
\label{tab:dap} \vspace{-8mm}
\end{table}

\begin{figure*}[t]
  \centering
  	\subfigure[AWA (FV=4K)] {
	\resizebox{0.23\linewidth}{!}{\includegraphics[trim=35 5 75 35, clip=true]{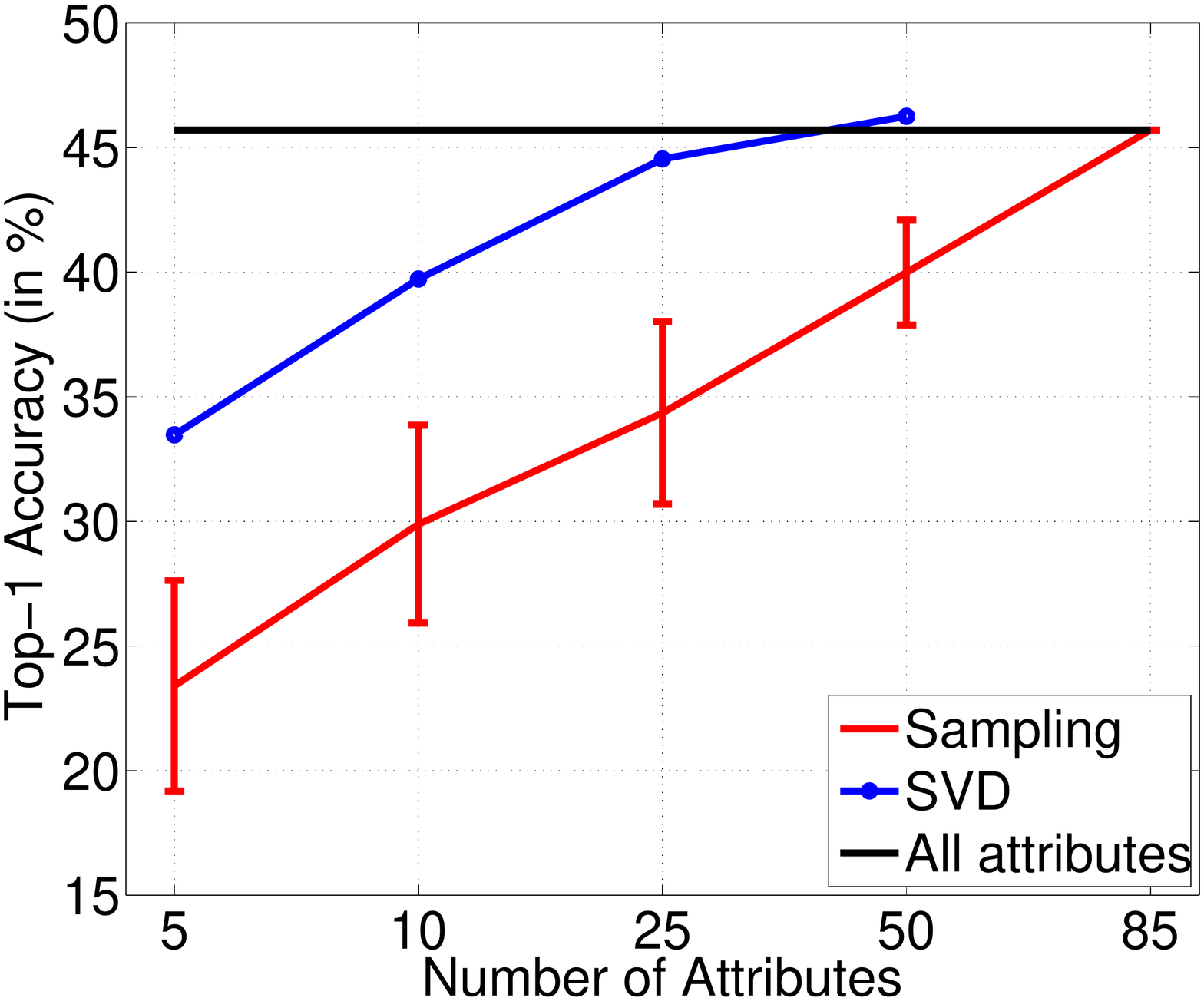}}
	\label{fig:svd_AWA_4K}
	}
	\subfigure[CUB (FV=4K)] {
  	\resizebox{0.23\linewidth}{!}{\includegraphics[trim=35 5 75 20, clip=true]{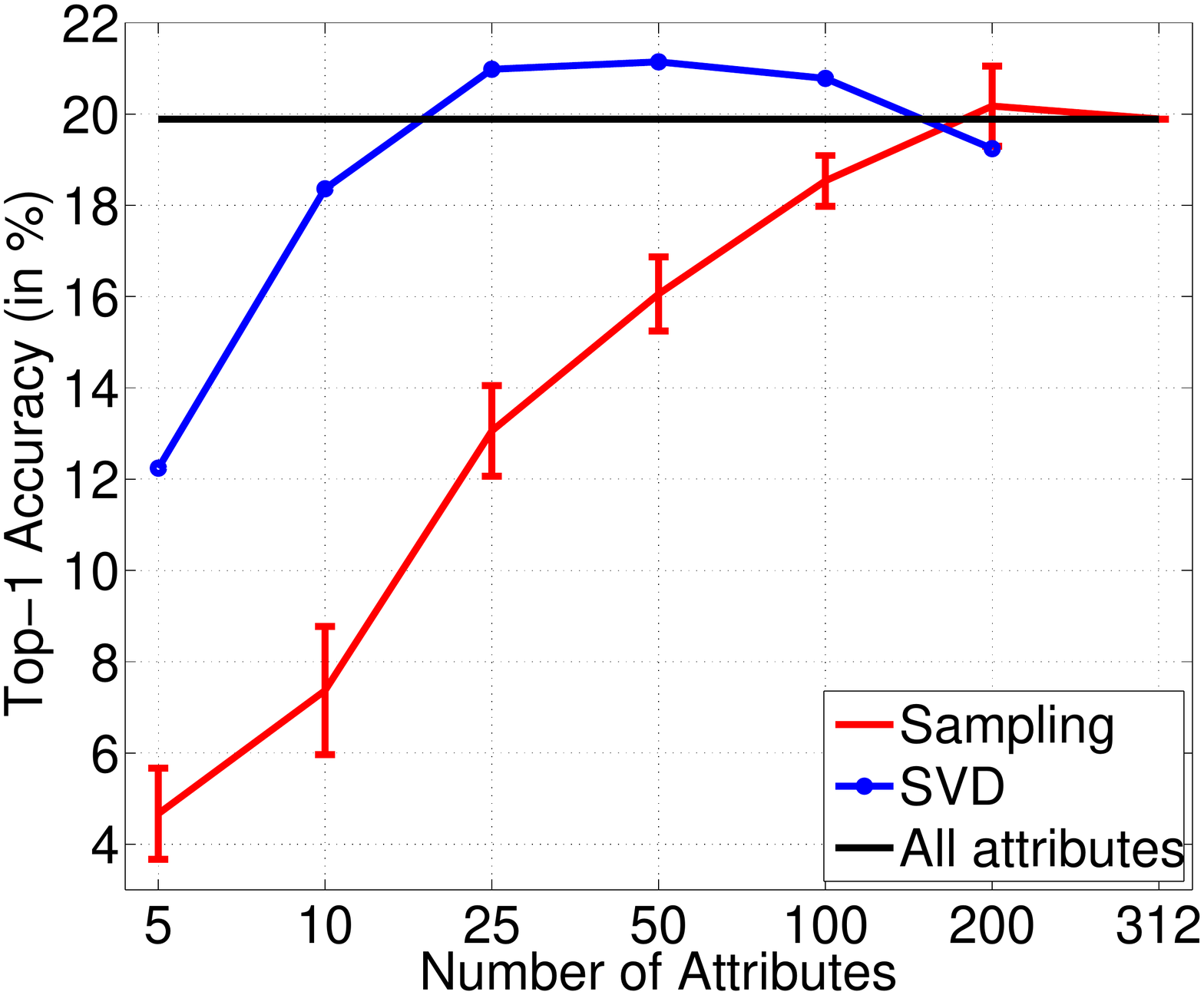}}
  		\label{fig:svd_CUB_4K}
	}
	\subfigure[AWA (FV=64K)] {
	\resizebox{0.23\linewidth}{!}{\includegraphics[trim=35 5 75 20, clip=true]{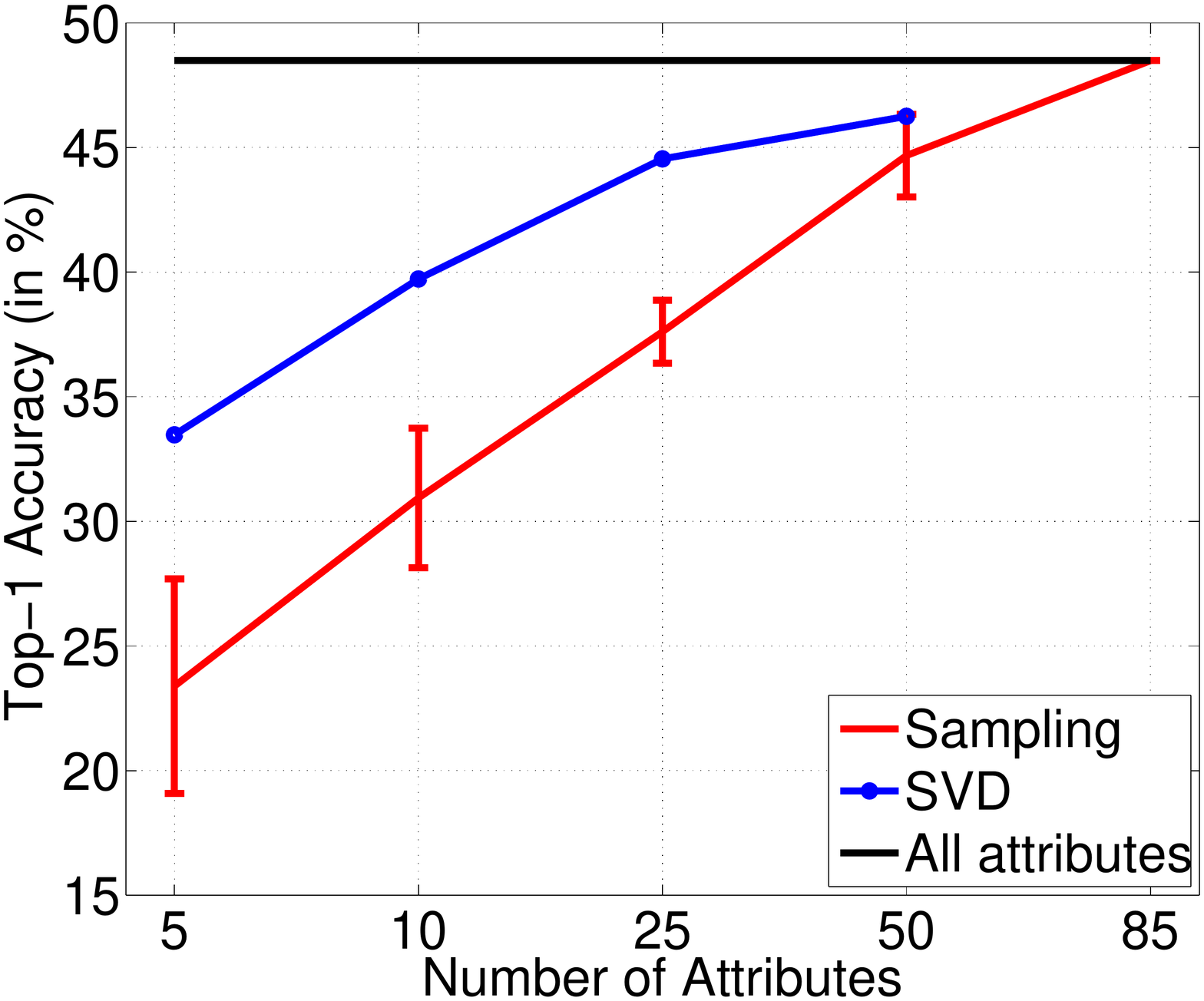}}
		\label{fig:svd_AWA_64K}
	}
	\subfigure[CUB (FV=64K)] {
	\resizebox{0.23\linewidth}{!}{\includegraphics[trim=35 5 75 20, clip=true]{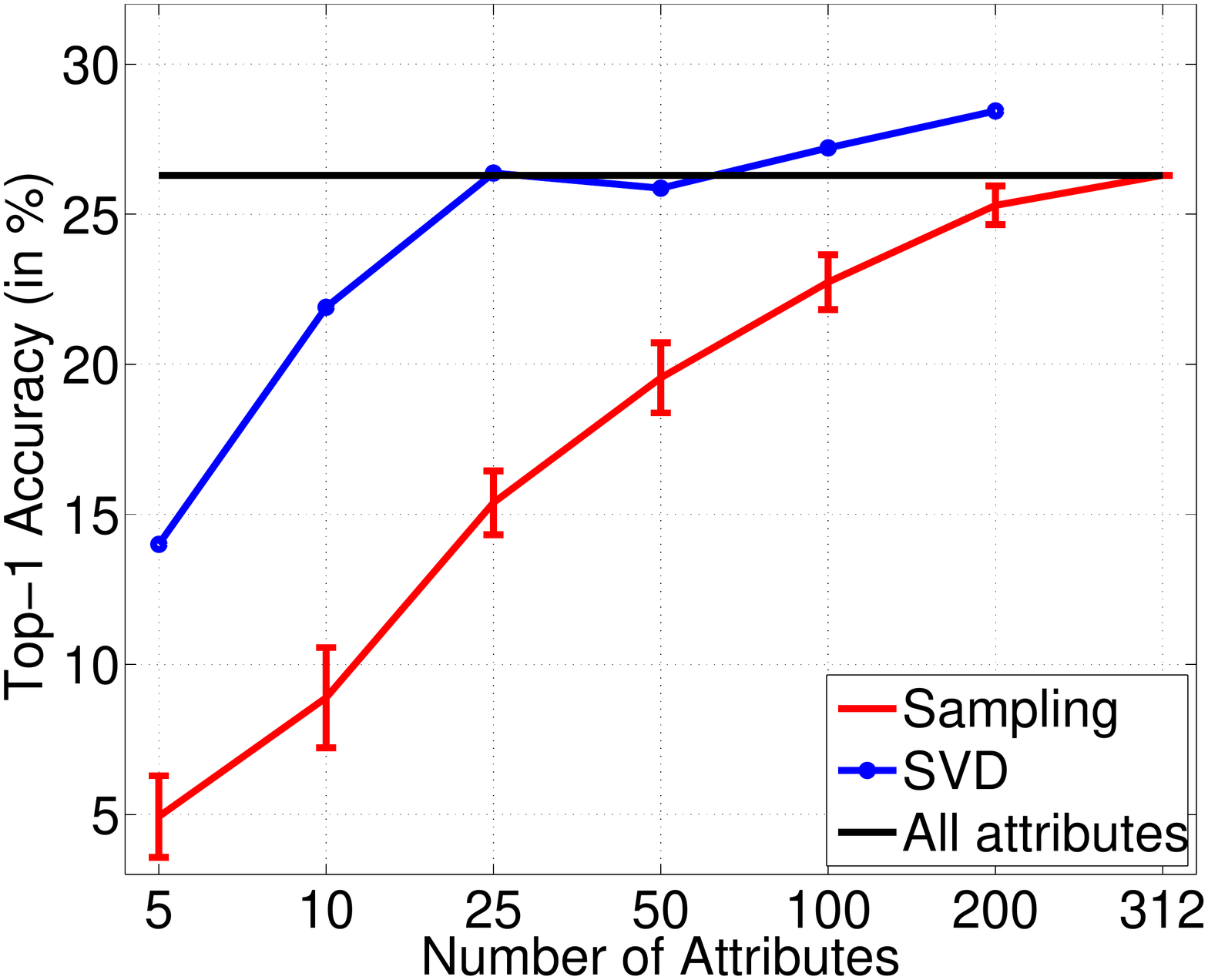}}
		\label{fig:svd_CUB_64K}
  	}
  \caption{Classification accuracy on AWA and CUB as a function of the label embedding dimensionality. We compare the baseline which uses all attributes, with an SVD dimensionality reduction and a sampling of attributes (we report the mean and standard deviation over 10 samplings).}
\label{fig:svd} \vspace{-4mm}
\end{figure*}

\vspace{2mm}\noindent
{\bf Comparison with DAP.} In this section we compare our approach to direct attribute prediction (DAP)~\cite{LNH09}. We start by giving a short description of DAP and, then, present the results of the
comparison. 

In DAP, an image $x$ is assigned to the class $y$, which has the highest posterior probability:
\begin{equation}
p(y|x) \propto \prod_{e=1}^E p(a_e = \rho_{y,e}|x) .
\end{equation}
$\rho_{y,e}$ is the binary association measure between attribute $a_e$ and class $y$.
$p(a_e=1|x)$ is the probability that image $x$ contains attribute
$e$.
We train for each attribute one linear classifier on the FVs. We use a (regularized) logistic loss which provides an attribute classification accuracy similar to SVM but with the added benefit that its output is already a probability.

Table \ref{tab:dap}(left) compares the proposed ALE to DAP for 64K-dim FVs. Our implementation of DAP obtains 41.0\% accuracy on AWA and 12.3\% on CUB. Our result for DAP on AWA is comparable to the 40.5\% accuracy reported by Lampert. Note however that the features are different. Lampert uses bag-of-features and a non-linear kernel classifier ($\chi^2$ SVMs), whereas we use Fisher vectors and a linear SVM. Linear SVMs enable us to run experiments more efficiently. We observe that on both datasets, the proposed ALE outperforms DAP significantly: 48.5\% \vs 41.0\% top-1 accuracy on AWA and 26.9\% \vs 12.3\% on CUB.

\vspace{2mm}\noindent
{\bf Attribute Correlation.}
While correlation in the input space is a well-studied topic, comparatively little work has been done to measure the correlation in the output space. Here, we reduce the output space dimensionality and study the impact on the classification accuracy. It is worth noting that reducing the output dimensionality leads to significant speed-ups  at training and test times. We explore two different techniques: Singular Value Decomposition (SVD) and attribute sampling. We learn the SVD on AWA (resp. CUB) on the 50$\times$85 (resp. 200$\times$312) $\Phi^{\cal A}$ matrix. For the sampling, we sub-sample a fixed number of attributes and repeat the experiments 10 times for 10 different random sub-samplings. The results of these experiments are presented in Figure~\ref{fig:svd}. 

We can conclude that there is a significant amount of correlation between attributes. For instance, on AWA with 4K-dim FVs (Figure \ref{fig:svd_AWA_4K}) when reducing the output dimensionality to 25, 
we lose less than 2\% accuracy and with a reduced dimensionality of 50, we perform
even slightly better than using all the attributes. On the same dataset with 64K-dim FVs (Figure \ref{fig:svd_AWA_64K}) 
the accuracy drops from 48.5\% to approximately 45\% when reducing from an 85-dim space to a 25-dim space.
More impressively, on CUB with 4K-dim FVs (Figure \ref{fig:svd_CUB_4K}) with a reduced dimensionality to 25, 50 or 100 from 312, 
the accuracy is better than the configuration that uses all the attributes. On the same dataset with 64K-dim FVs
 (Figure \ref{fig:svd_CUB_64K}), with 25 dimensions the accuracy is on par with the 312-dim embedding.
SVD outperforms a random sampling of the attribute dimensions, although there is no guarantee that
SVD will select the most informative dimensions (see for instance the small pit in performance
on CUB at 50 dimensions). In random sampling of output embeddings, the choice of the attributes
seems to be an important factor that affects the descriptive power of output embeddings.
Consequently, the variance is higher (\eg see Figures \ref{fig:svd_AWA_4K} and Figure \ref{fig:svd_AWA_64K} 
with a reduced attribute dimensionality of 5 or 10) when a small number of attributes is selected.
In the following experiments, we do not use dimensionality reduction of the attribute embeddings.

\begin{table}[t]
 \begin{center}
  \small
  \begin{tabular}{|r|c|c|c|c|c|c|}
\hline
& ALE & HLE & WLE & \specialcell{AHLE \\ early} & \specialcell{AHLE \\late} \\
\hline
AWA & 48.5 & 40.4 & 32.5 & 46.8 & \bf{49.4} \\
\hline
CUB & 26.9 & 18.5 & 16.8 & 27.1 & \bf{27.3} \\
\hline
\end{tabular}
\end{center}
\caption{Comparison of attributes (ALE), hierarchies (HLE) and Word2Vec (WLE) for label embedding. We consider the combination of ALE and HLE by simple concatenation (AHLE early) or by the averaging of the scores (AHLE late). We use 64K FVs.}
\label{tab:hie} \vspace{-7mm}
\end{table}

\vspace{2mm}\noindent
{\bf Attribute interpretability.}
In ALE, each column of $W$ can be interpreted as an attribute classifier and $\theta(x)'W$ as a vector of attribute scores of $x$. However, one major difference with DAP is that we do not optimize for attribute classification accuracy. This might be viewed as a disadvantage of our approach as we might loose interpretability, an important property of attribute-based systems when, for instance, one wants to include a human in the loop \cite{BWB10,WBPB11}. We, therefore, measured the attribute prediction accuracy of DAP and ALE. For each attribute, following \cite{LNH09}, we measure the AUC on the set of the evaluation classes and report the mean. 

\begin{figure*}
	\centering
\includegraphics[width=\textwidth]{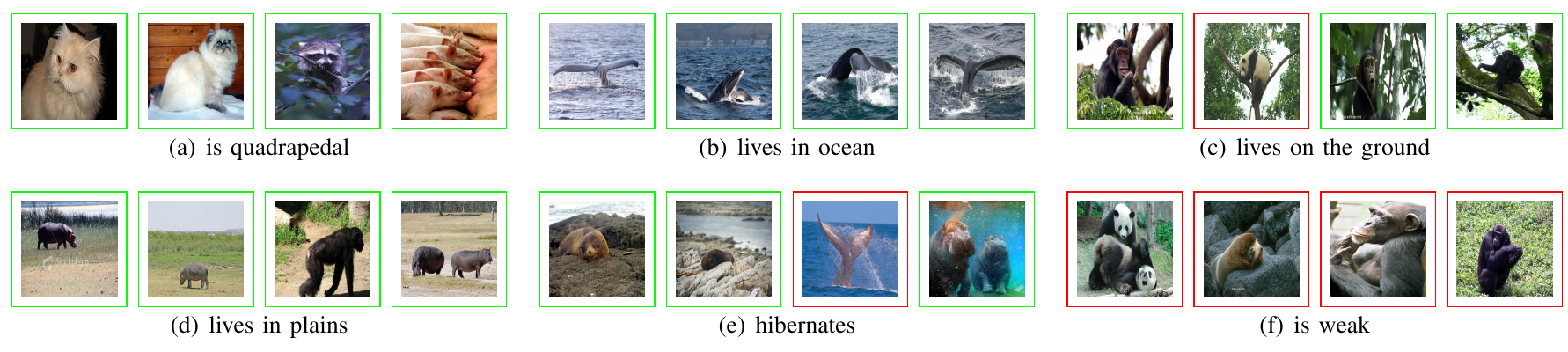}

	\caption{Sample attributes recognized with high ($>$ 90\%) accuracy (top) and low (\ie $<$50\%) accuracy (bottom) by ALE on AWA.
          For each attribute we show the images ranked highest. 
          Note that a AUC $<$ 50\% means that the prediction is worse than random on average. 
          The images whose attribute is predicted correctly are circled in green and those whose attribute is predicted incorrectly 
          are circled in red.}
\label{fig:awa_att} \vspace{-5mm}
\end{figure*}

Attribute prediction scores are shown in Table \ref{tab:dap}(right). On AWA, the DAP and ALE methods obtain the same AUC accuracy of 72.7\%. On the other hand, on CUB the DAP method obtains 64.8\% AUC whereas ALE is 5.4\% lower with 59.4\% AUC. As a summary, the attribute prediction accuracy of DAP is at least as high as that of ALE. This is expected since DAP optimizes directly attribute-classification accuracy. However, the AUC for ALE is still reasonable, especially on AWA (performance is on par). Thus, our learned attribute classifiers should still be interpretable. We provide qualitative results on AWA in Figure~\ref{fig:awa_att}: we show the four highest ranked images for some of the attributes with the highest AUC scores (namely $>$90\%) and lowest AUC scores (namely $<$50\%). 

\vspace{2mm}\noindent
{\bf Comparison of ALE, HLE and WLE.} We now compare different sources of side information. Results are provided in Table \ref{tab:hie}. On AWA, ALE obtains 48.5\% accuracy, HLE obtains 40.4\% and WLE obtains 32\% accuracy. On CUB, ALE obtains 26.9\% accuracy, HLE obtains 18.5\% and WLE obtains 16.8\% accuracy. Note that in \cite{APHS13}, we reported better results on AWA with HLE compared to ALE. The main difference with the current experiment is that we use continuous attribute encodings while \cite{APHS13} was using a binary encoding. Note also that the comparatively poor performance of WLE with respect to ALE and HLE is not unexpected: while ALE and HLE rely on strong expert supervision, WLE is computed in an unsupervised manner from Wikipedia.

We also consider the combination of attributes and hierarchies (we do not consider the combination of WLE with other embeddings given its relatively poor performance). We explore two simple alternatives: the concatenation of the embeddings (AHLE early) and the late fusion of classification scores calculated by averaging the scores obtained using ALE and HLE separately (AHLE late). On both datasets, late fusion has a slight edge over early fusion and leads to a small improvement over ALE alone (+0.9\% on AWA and +0.4\% on CUB).

In what follows, we do not report further results with WLE given its relatively poor performance and focus on ALE and HLE.

\vspace{2mm} \noindent
{\bf Comparison with the state-of-the-art.}
We can compare our results to those published in the literature on AWA since we are using the standard training/testing protocol (there is no such zero-shot protocol on CUB). To the best of our knowledge, the best zero-shot recognition results on AWA are those of Yu \etal~\cite{YCFSC13} with 48.3\% accuracy. We report 48.5\% with ALE and 49.4\% with AHLE (late fusion of ALE and HLE). Note that we use different features.

\subsection{Few-Shots Learning}
\label{sec:few}

\begin{figure}[t]
  \centering
 	\subfigure[AWA (FV=64K)] {
	\resizebox{0.7\linewidth}{!}{\includegraphics[trim=35 5 75 20, clip=true]{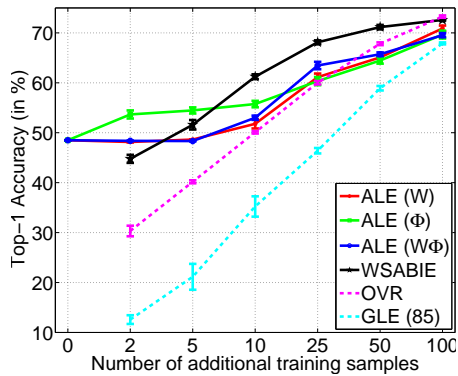}}
	\label{fig:few_AWA_256}
	}
	\subfigure[CUB (FV=64K)] {
	\resizebox{0.7\linewidth}{!}{\includegraphics[trim=35 5 75 20, clip=true]{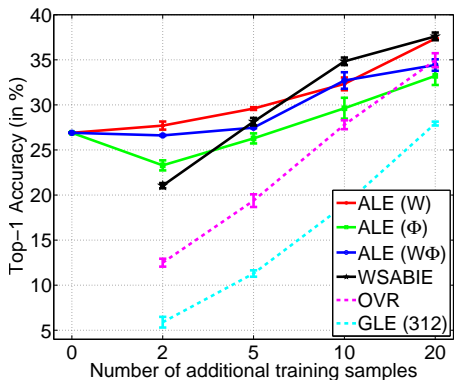}}
	\label{fig:few_CUB_256}
	}
  \caption{Classification accuracy on AWA and CUB as a function of the number of training samples per class. To train the classifiers, we use all the images of the training ``background'' classes (used in zero-shot learning), and a small number of images randomly drawn from the relevant evaluation classes. Reported results are 10-way in AWA and 50-way in CUB.}
  \label{fig:few} \vspace{-5mm}
\end{figure}

\noindent {\bf Set-up.}
In these experiments, we assume that we have few (\eg 2, 5, 10, etc.) training samples for a set of classes of interest (the 10 AWA and 50 CUB evaluation classes) in addition to all the samples from a set of ``background classes'' (the remaining 40 AWA and 150 CUB classes). For each evaluation class, we use approximately half of the images for training (the 2, 5, 10, etc. training samples are drawn from this pool) and the other half for testing. The minimum number of images per class in the evaluation set is 302 (AWA) and 42 (CUB). To have the same number of training samples, we use 100 images (AWA) and 20 images (CUB) per class as training set and the remaining images for testing. 

\vspace{2mm} \noindent
{\bf Algorithms.} We compare the proposed ALE with three baselines: OVR, GLE and WSABIE. We are especially interested in analyzing the following factors: (i) the influence of parameter sharing (ALE, GLE, WSABIE) \vs no parameter sharing (OVR), (ii) the influence of learning the embedding (WSABIE) \vs having a fixed embedding (ALE, OVR and GLE), and (iii) the influence of prior information (ALE) \vs no prior information (OVR, GLE and WSABIE) 

For ALE and WSABIE, $W$ is initialized to the matrix learned in the zero-shot experiments. For ALE, we experimented with three different learning variations:
\begin{itemize}
\item ALE($W$) consists in learning the parameters $W$ and keeping the embedding fixed ($\Phi = \Phi^{\cal A}$).
\item ALE($\Phi$) consists in learning the embedding parameters $\Phi$ and keeping $W$ fixed.
\item ALE($W\Phi$) consists in learning both $W$ and $\Phi$.
\end{itemize}

While both ALE($W$) and ALE($\Phi$) are implemented by stochastic (sub)gradient descent (see Algorithm~\ref{alg:sgd} in Sec.~\ref{sec:obj}), ALE($W\Phi$) is implemented by stochastic alternating optimization. Stochastic alternating optimization alternates between SGD for optimizing over the variable $W$ and optimizing over the variable $\Phi$. Theoretical convergence of SGD for ALE($W$) and ALE($\Phi$) follows from standard results in stochastic optimization with convex non-smooth objectives~\cite{SSSC11,ShaiShai:2014}. Theoretical convergence of the stochastic alternating optimization is beyond the scope of the paper. Experimental results show that the strategy actually works fine empirically.

\vspace{2mm} \noindent
{\bf Results.} We show results in Figure~\ref{fig:few} for AWA and CUB using 64K-dim features. We can draw the following conclusions. First, GLE underperforms all other approaches for limited training data which shows that random embeddings are not appropriate in this setting. Second, in general, WSABIE and ALE outperform OVR and GLE for small training sets (\eg for less than 10 training samples) which shows that learned embeddings (WSABIE) or embeddings based on prior information (ALE) can be effective when training data is scarce. Third, for tiny amounts of training data (\eg 2-5 training samples per class), ALE outperforms WSABIE which shows the importance of prior information in this setting. Fourth, all variations of ALE -- ALE($W$), ALE($\Phi$) and ALE($W\Phi$) -- perform somewhat similarly. Fifth, as the number of training samples increases, all algorithms seem to converge to a similar accuracy, \ie as expected parameter sharing and prior information are less crucial when training data is plentiful.

\subsection{Learning and testing on the full datasets}
\label{sec:full}
In these experiments, we learn and test the classifiers on the 50 AWA (resp. 200 CUB) classes. For each class, we reserve approximately half of the data for training and cross-validation purposes and half of the data for test purposes. On CUB, we use the standard training/test partition provided with the dataset. Since the experimental protocol in this section is significantly different from the one chosen
for zero-shot and few-shots learning, the results cannot be directly compared with those of the previous sections.

\vspace{2mm}
\noindent {\bf Comparison of output encodings.}
We first compare different encoding techniques (continuous embedding \vs binary embedding) and normalization strategies (with/without mean centering and with/without $\ell_2$-normalization). The results are provided in Table \ref{tab:disc_cont_all}. We can draw the following conclusions.

As is the case for zero-shot learning, mean-centering has little impact and $\ell_2$-normalization consistently improves performance, showing the importance of normalized outputs. On the other hand, a major difference with the zero-shot case is that the $\{0,1\}$ and continuous embeddings perform on par. On AWA, in the 64K-dim FVs case, ALE with continuous embeddings leads to 53.3\% accuracy whereas $\{0,1\}$ embeddings leads to 52.5\% (0.8\% difference). On CUB with 64K-dim FVs, ALE with continuous embeddings leads to 21.6\% accuracy while $\{0,1\}$ embeddings lead to 21.4\% (0.2\% difference).
This seems to indicate that the quality of the prior information used to perform label embedding has less impact when training data is plentiful. 

\begin{table}[t]
 \begin{center}
  \small
  \begin{tabular}{|r|r|c|c|c|c|}
  \hline
  & & \multicolumn{4}{c|}{AWA}\\
\hline
& & \multicolumn{2}{c|}{FV=4K} & \multicolumn{2}{c|}{FV=64K} \\
\hline
$\mu$ & $\ell_2$ & $\{0,1\}$ & cont & $\{0,1\}$ & cont \\
\hline
no & no & 42.3 & 41.6 & 45.3 & 46.2 \\
\hline
no & yes & 44.3 & 44.6 & 52.5 & {\bf 53.3} \\
\hline
yes & no & 42.2 & 41.6 & 45.8 & 46.2 \\
\hline
yes & yes & {\bf 44.8} & 44.5 & 51.3 & 52.0 \\
\hline
\hline
  & & \multicolumn{4}{c|}{CUB}\\
\hline
& & \multicolumn{2}{c|}{FV=4K} & \multicolumn{2}{c|}{FV=64K} \\
\hline
$\mu$ & $\ell_2$ & $\{0,1\}$ & cont & $\{0,1\}$ & cont \\
\hline
no & no & 13.0 & 13.9 & 16.5 & 16.7 \\
\hline
no & yes & 16.2 & {\bf 17.5} & 21.4 & {\bf 21.6} \\
\hline
yes & no & 13.2 & 13.9 & 16.5 & 16.7 \\
\hline
yes & yes & 16.1 & 17.3 & 17.3 & {\bf 21.6} \\
\hline
\end{tabular}
\end{center}
\caption{Comparison of different output encodings:
binary $\{0,1\}$ encoding, continuous encoding, 
with/without mean-centering ($\mu$) and with/without $\ell_2$-normalization }
\label{tab:disc_cont_all} \vspace{-7mm}
\end{table}

\vspace{2mm}
\noindent {\bf Comparison of output embedding methods.} We now compare on the full training sets several learning algorithms: OVR, GLE with a costly setting $E=2,500$ output dimensions this was the largest output dimensionality allowing us to run the experiments in a reasonable amount of time), WSABIE (with cross-validated $E$), ALE (we use the ALE($W$) variant where the embedding parameters are kept fixed), HLE and AHLE (with early and late fusion). Results are provided in Table \ref{tab:full}. 

We can observe that, in this setting, all methods perform somewhat similarly. Especially, the simple OVR and GLE baselines provide a competitive performance: OVR outperforms all other methods on CUB and GLE performs best on AWA. This confirms that the quality of the embedding has little importance when training data is plentiful.

\begin{table}[t]
 \begin{center}
  \small
  \resizebox{\linewidth}{!}{
  \begin{tabular}{|r|c|c|c|c|c|c|c|c| }
\hline
 & OVR & GLE & \footnotesize{WSABIE} & ALE & HLE & \specialcell{AHLE \\early} & \specialcell{AHLE \\ late} \\
\hline
AWA & 52.3 & \textbf{56.1} & 51.6 & 52.5 & 55.9 & 55.3 & 55.8\\
\hline
CUB & \bf{26.6} & 22.5 & 19.5 & 21.6 & 22.5 & 24.6 & 25.5 \\
\hline
\end{tabular}
}
\end{center}
\caption{Comparison of different output embedding methods (OVR, GLE, WSABIE, ALE, HLE, AHLE early and AHLE late ) on the full AWA and CUB datasets (resp. 50 and 200 classes). We use 64K FVs.
}
\label{tab:full} \vspace{-7mm}
\end{table}

\begin{figure}[t]
  \centering
	\subfigure[AWA (FV=64K)] {
	\resizebox{0.7\linewidth}{!}{\includegraphics[trim=35 5 50 20, clip=true]{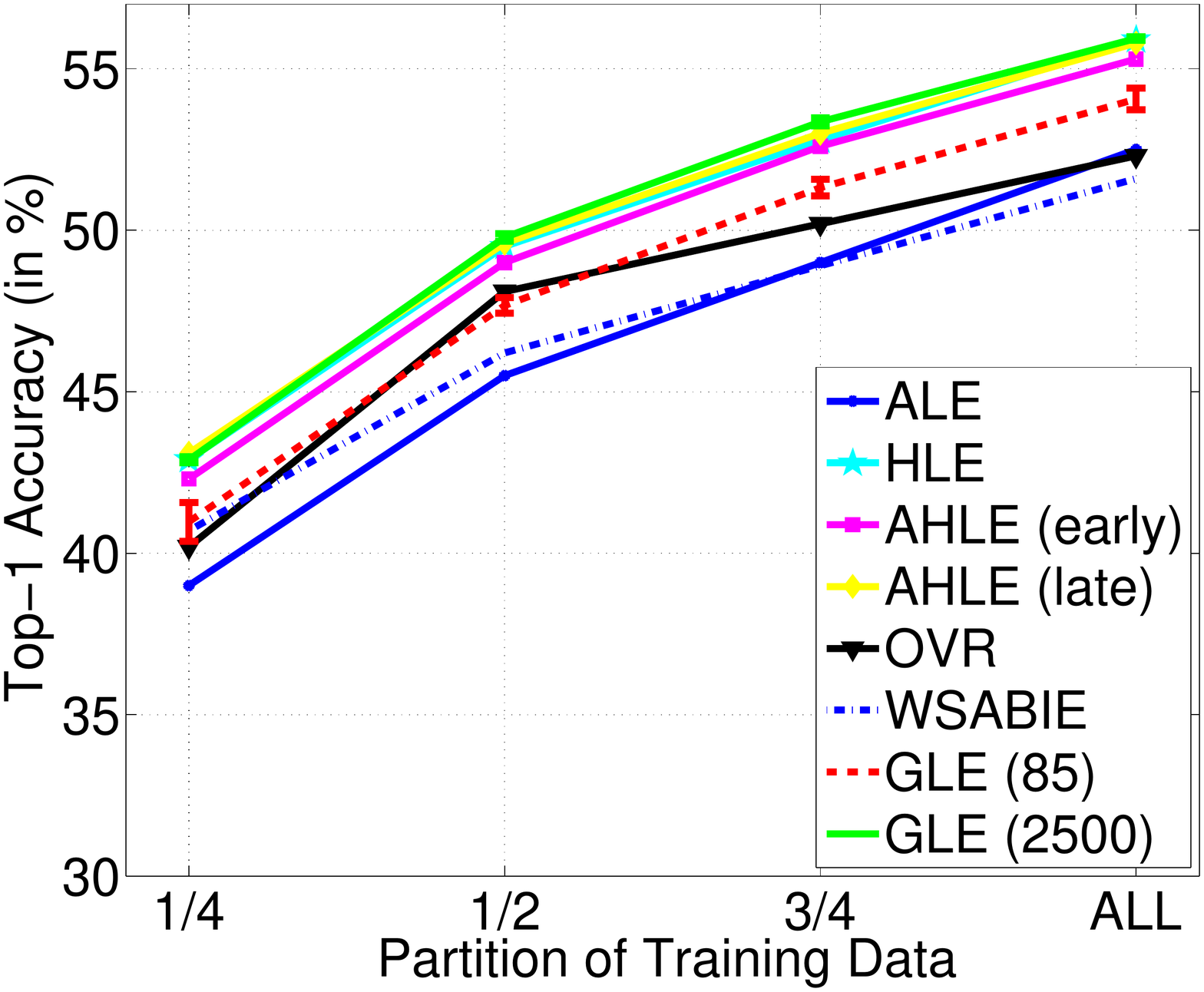}}
	}
	\subfigure[CUB (FV=64K)] {
	\resizebox{0.7\linewidth}{!}{\includegraphics[trim=35 5 50 20, clip=true]{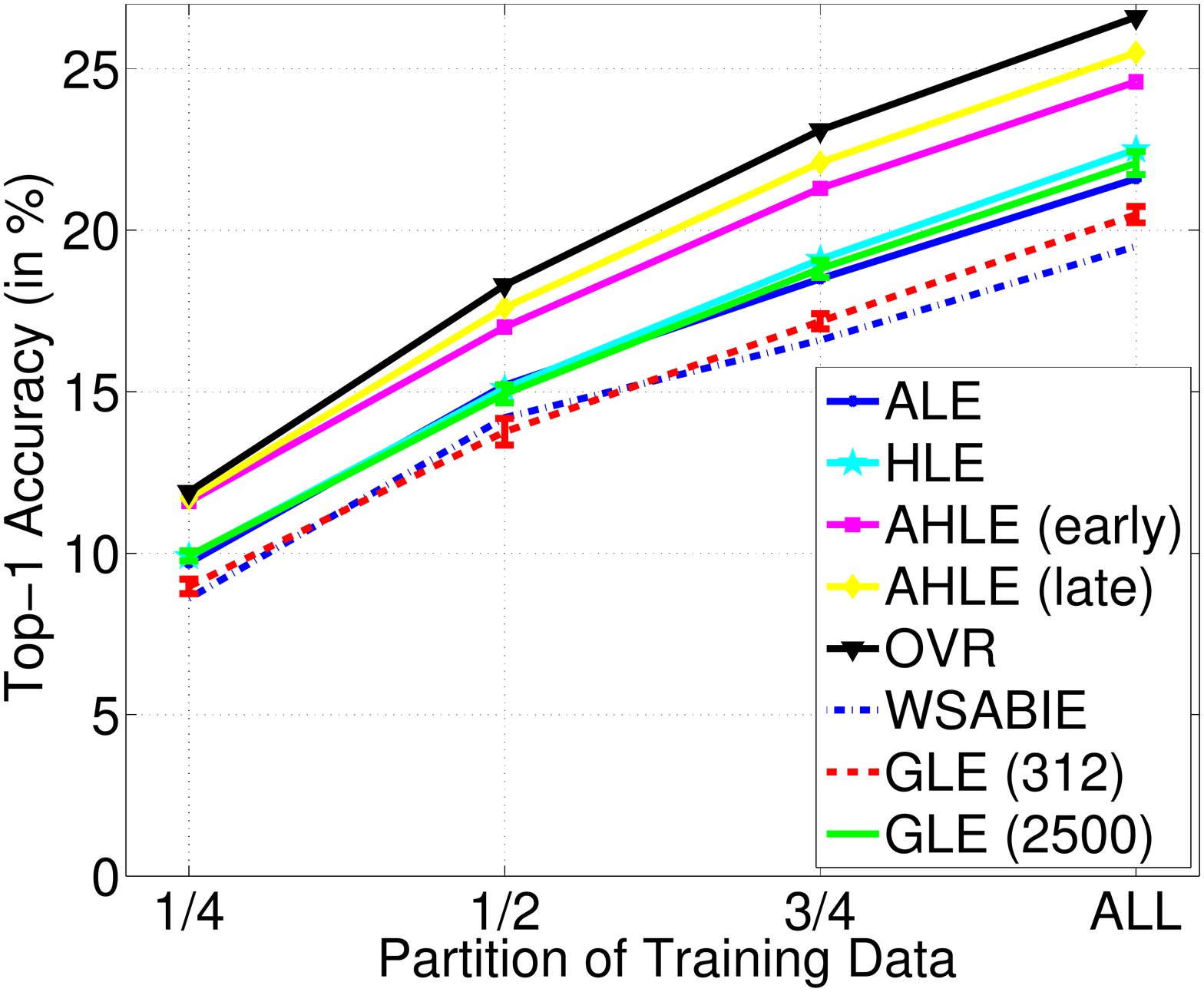}}
	}
  \caption{Learning on AWA and CUB using 1/4, 1/2, 3/4 and all the training data. Compared output embeddings: OVR, GLE, WSABIE, ALE, HLE, AHLE early and AHLE late. Experiments repeated 10 times for different sampling of Gaussians. We use 64K FVs.
}
  \label{fig:part_seed} \vspace{-7mm}
\end{figure}

\vspace{2mm}
\noindent {\bf Reducing the training set size.} We also studied the effect of reducing the amount of training data by using only 1/4, 1/2 and 3/4 of the full training set. We therefore sampled the corresponding fraction of images from the full training set and repeated the experiments ten times with ten different samples. For these experiments, we report GLE results with two settings:
using a low-cost setting, \ie using the same number of output dimensions $E$ as ALE (\ie 85 for AWA and 312 for CUB) and using a high-cost setting, \ie using a large number of output dimensions ($E=2,500$ -- see comment above about the choice of the value $2,500$). We show results in Figure \ref{fig:part_seed}.

On AWA, GLE outperforms all alternatives, closely followed by AHLE late. On CUB, OVR outperforms all alternatives, closely followed again by AHLE late. ALE, HLE and GLE with high-dimensional embeddings perform similarly. For these experiments, a general conclusion is that, when we use high dimensional features, even simple algorithms such as the OVR which are not well-justified for multi-class classification problems can lead to state-of-the-art performance.

\section{Conclusion}
We proposed to cast the problem of attribute-based classification as one of label-embedding. The proposed Attribute Label Embedding (ALE) addresses in a principled fashion the limitations of the original DAP model. First, we solve directly the problem at hand (image classification) without introducing an intermediate problem (attribute classification). Second, our model can leverage labeled training data (if available) to update the label embedding, using the attribute embedding as a prior. Third, the label embedding framework is not restricted to attributes and can accommodate other sources of side information such as class hierarchies or words embeddings derived from textual corpora.

In the zero-shot setting, we improved image classification results with respect to DAP without losing attribute interpretability. Continuous attributes can be effortlessly used in ALE, leading to a large boost in zero-shot classification accuracy. As an addition, we have shown that the dimensionality of the output space can be significantly reduced with a small loss of accuracy. In the few-shots setting, we showed improvements with respect to the WSABIE algorithm, which learns the label embedding from labeled data but does not leverage prior information.

Another important contribution of this work was to relate different approaches to label embedding: data-independent approaches ({\em e.g.} OVR, GLE), data-driven approaches ({\em e.g.} WSABIE) and approaches based on side information ({\em e.g.} ALE, HLE and WLE). 
We present here a unified framework allowing to compare them in a systematic manner.

Learning to combine several inputs has been extensively studied in machine learning and computer vision, whereas learning to combine outputs is still largely unexplored. We believe that it is a worthwhile research path to pursue.

\section*{Acknowledgments}
{\small The Computer Vision group at XRCE is funded partially by the Project Fire-ID (ANR-12-CORD-0016).
The LEAR team of Inria is partially funded by ERC Allegro, and European integrated project AXES. }

\bibliographystyle{ieee}
\bibliography{TPAMI}

\begin{IEEEbiography}[{\includegraphics[width=1in,height=1.2in,clip,keepaspectratio]{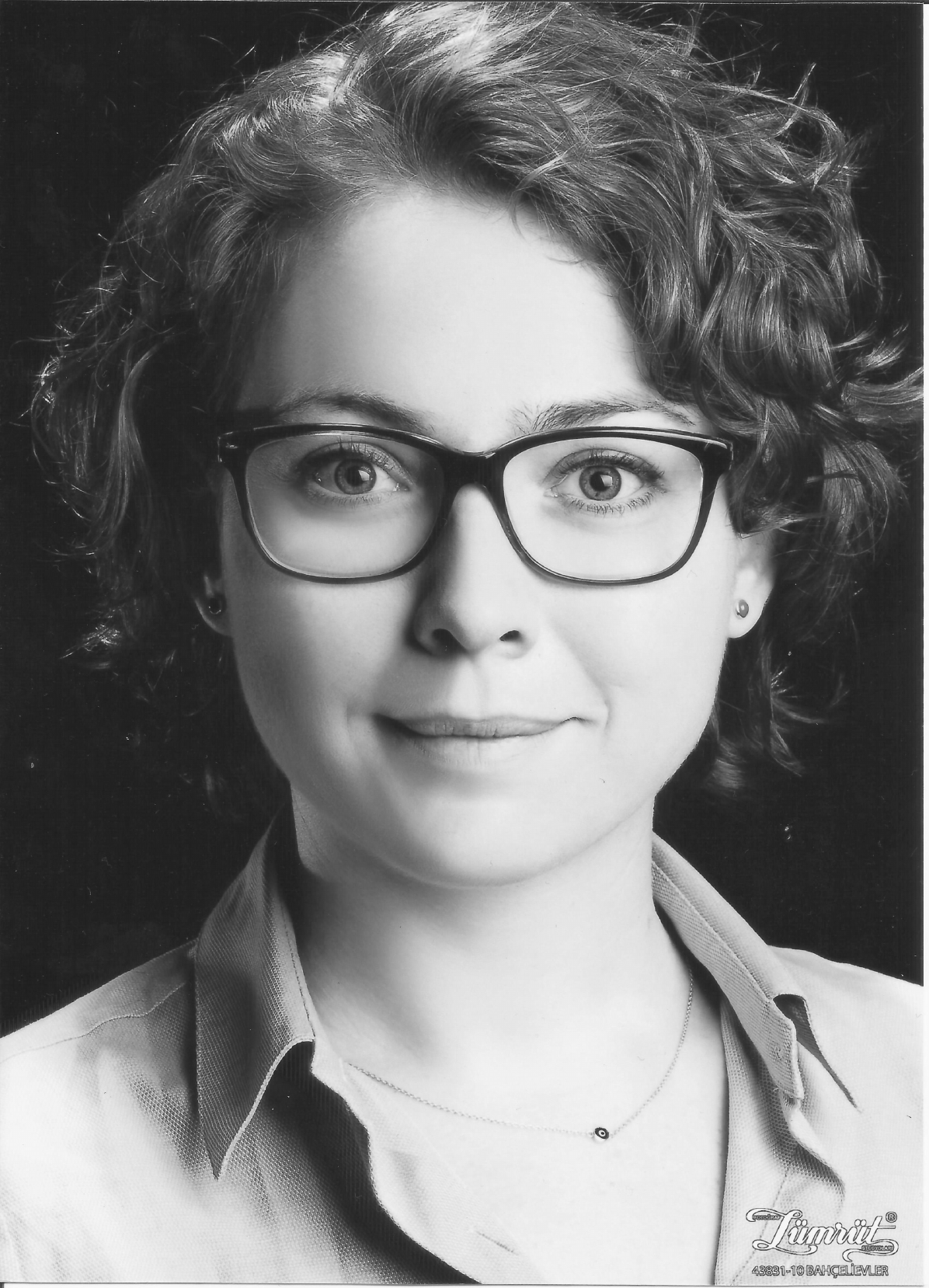}}]{Zeynep Akata}
received her MSc degree from RWTH Aachen and her PhD degree from Universit\'e de Grenoble within a collaboration between
XRCE and INRIA. In 2014, she received Lise-Meitner Award for Excellent Women in Computer Science and 
currently working as a post-doctoral researcher in Max Planck Institute of Informatics in Germany. 
Her research interests include machine learning methods for large-scale and fine-grained image classification. 
\end{IEEEbiography} \vspace{-10mm}

\begin{IEEEbiography}[{\includegraphics[width=1in,height=1.2in,clip,keepaspectratio]{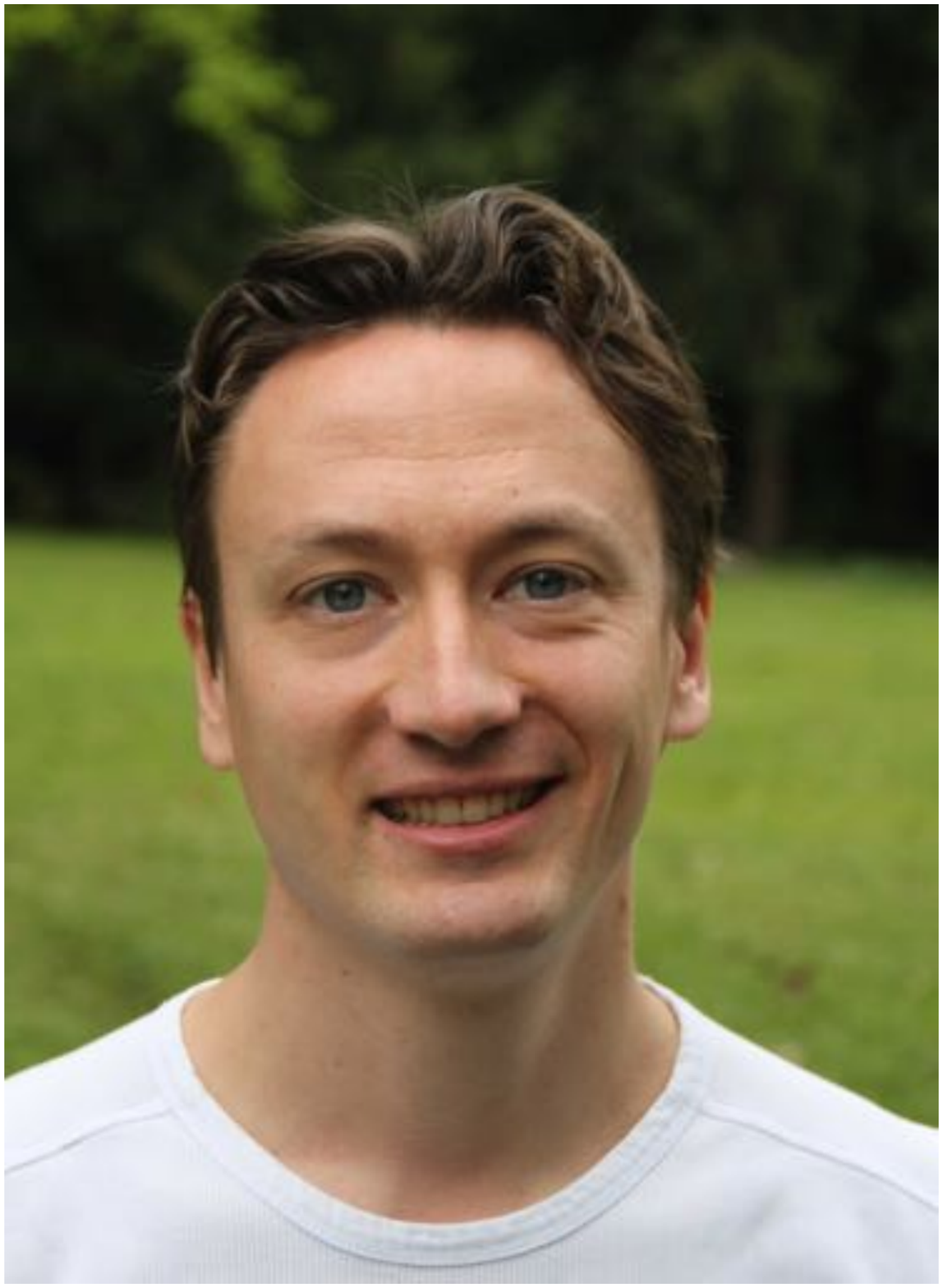}}]{Florent Perronnin}
holds an Engineering degree from the Ecole Nationale 
Sup\'erieure des T\'el\'ecommunications and a Ph.D. degree 
from the Ecole Polytechnique F\'ed\'erale de Lausanne.
In 2005, he joined the Xerox Research Centre Europe in Grenoble where he currently manages the Computer Vision team.
His main interests are in the application of machine learning to computer vision tasks such as image classification, retrieval
or segmentation.
\end{IEEEbiography} \vspace{-10mm}

\begin{IEEEbiography}[{\includegraphics[width=1in,height=1.2in,clip,keepaspectratio]{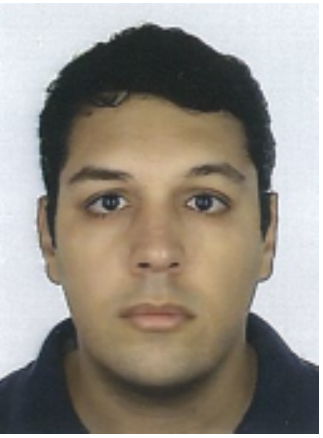}}]{Zaid Harchaoui}
graduated from the Ecol\'e Nationale Superieure des Mines, Saint-Etienne, France, in 2004, and the Ph.D. degree 
from ParisTech, Paris, France. Since 2010, he is a permanent researcher in the LEAR team, INRIA Grenoble, France. 
His research interests include statistical machine learning, kernel-based methods, signal processing, and computer vision.
\end{IEEEbiography} \vspace{-10mm}

\begin{IEEEbiography}[{\includegraphics[width=1in,height=1.2in,clip,keepaspectratio]{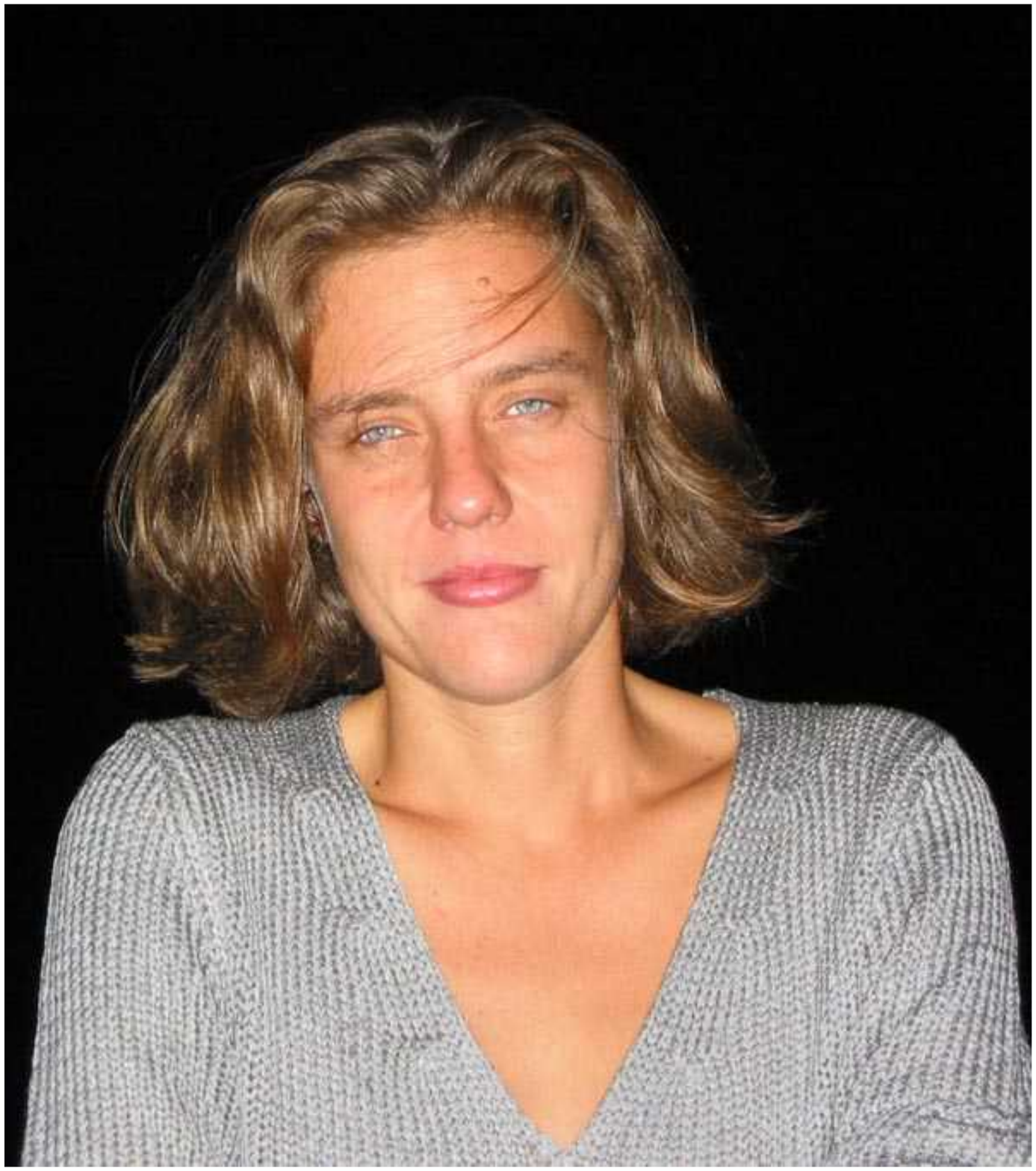}}]{Cordelia Schmid}
holds a M.S. degree in computer science from the
University of Karlsruhe and a doctorate from the Institut National
Polytechnique de Grenoble. She is a research director at INRIA
Grenoble where she directs the LEAR team. She is the author of over a
hundred technical publications. In 2006 and 2014, she was awarded the
Longuet-Higgins prize for fundamental contributions in computer vision
that have withstood the test of time. In 2012, she obtained an ERC
advanced grant for "Active large-scale learning for visual
recognition". She is a fellow of IEEE.
\end{IEEEbiography}

\end{document}